\documentclass{IEEEtran}
\usepackage{cite}
\usepackage{amsmath,amssymb,amsfonts}
\usepackage{graphicx}
\usepackage{textcomp}
\usepackage{multirow}
\usepackage{hyperref}
\usepackage{multicol}
\usepackage{url}
\usepackage{booktabs}
\usepackage{makecell}
\usepackage{color}
\usepackage{bbding}
\usepackage{algorithm}
\usepackage{algpseudocode}
\usepackage{tikz}

\newcommand{\spm}{\mathbin{\raisebox{0.2ex}{\scalebox{0.65}{$\pm$}}}}

\def\BibTeX{{\rm B\kern-.05em{\sc i\kern-.025em b}\kern-.08em
		T\kern-.1667em\lower.7ex\hbox{E}\kern-.125emX}}
\ifCLASSINFOpdf
\else
\fi

\begin{document}
	\title{CmIVTP: Cross-modal Interaction-based Vessel Trajectory Prediction for Maritime Intelligence}
	\author{Yuxu Lu, Dong Yang, Xiaoyu Li, Mengwei Bao, and Congcong Zhao
		\thanks{This work was supported in part by the Research Grants Council of the Hong Kong Special Administrative Region under Grant PolyU 15225325. (Corresponding author: D. Yang)}
		\thanks{Y. Lu, X. Li, and C. Zhao are with the Department of Logistics and Maritime Studies, the Hong Kong Polytechnic University, Hong Kong 999077. (e-mail: yuxulouis.lu@connect.polyu.hk, xiaoyu0726.li@connect.polyu.hk, and cong-cong.zhao@connect.polyu.hk)}
        \thanks{D. Yang is with the Department of Logistics and Maritime Studies and the Research Centre for ESG Advancement (RCESGA), the Hong Kong Polytechnic University, Hong Kong 999077. (e-mail: dong.yang@polyu.edu.hk)}
		\thanks{M. Bao is with the Department of Logistics and Maritime Studies, the Hong Kong Polytechnic University, Hong Kong 999077, and also with the School of Navigation, Wuhan University of Technology, Wuhan 430063. (e-mail: mengweibao@whut.edu.cn)}
        }
\maketitle
\begin{abstract}
    Maritime intelligent transportation systems (MITS) are essential for ensuring navigation safety and efficiency in busy waterways. However, accurate vessel trajectory prediction remains challenging due to the limitations of single-source data. Automatic identification system (AIS) data is often sparse or unavailable for small vessels, while closed-circuit television (CCTV) data alone cannot fully capture dynamic vessel behavior. To mitigate these challenges, we propose a cross-modal interaction-based vessel trajectory prediction (named CmIVTP) framework to model the intricate interactions between vessel dynamics and environmental constraints. Specifically, we introduce a target-aware scene encoder to extract scene semantic features, effectively capturing vessel-environment interactions and enhancing trajectory prediction accuracy. In addition, we propose a cross-modal interaction transformer, which integrates AIS-derived motion features, CCTV-based environmental features, and scene representations. It leverages cross-modal attention mechanisms to simultaneously capture intra-modal semantics and inter-modal interactions, ensuring dynamically consistent and environmentally feasible predictions. Furthermore, we construct a vessel group trajectory bank by clustering historical AIS trajectories into representative motion patterns, providing an efficient and scalable approach for candidate trajectory generation. Additionally, we introduce the maritime multimodal dataset plus (named Maritime-MmD$^+$), a large-scale dataset that synchronizes AIS data and CCTV video data, providing robust support for multimodal trajectory prediction research. Extensive experiments demonstrate that CmIVTP achieves better performance on multimodal-driven vessel trajectory prediction benchmarks. The code resources for this work can be available at \url{https://github.com/LouisYxLu/CmIVTP}.
\end{abstract}

\section{Introduction}
\begin{figure}[t]
        \centering
        \setlength{\abovecaptionskip}{0.cm}
        \includegraphics[width=1.00\linewidth]{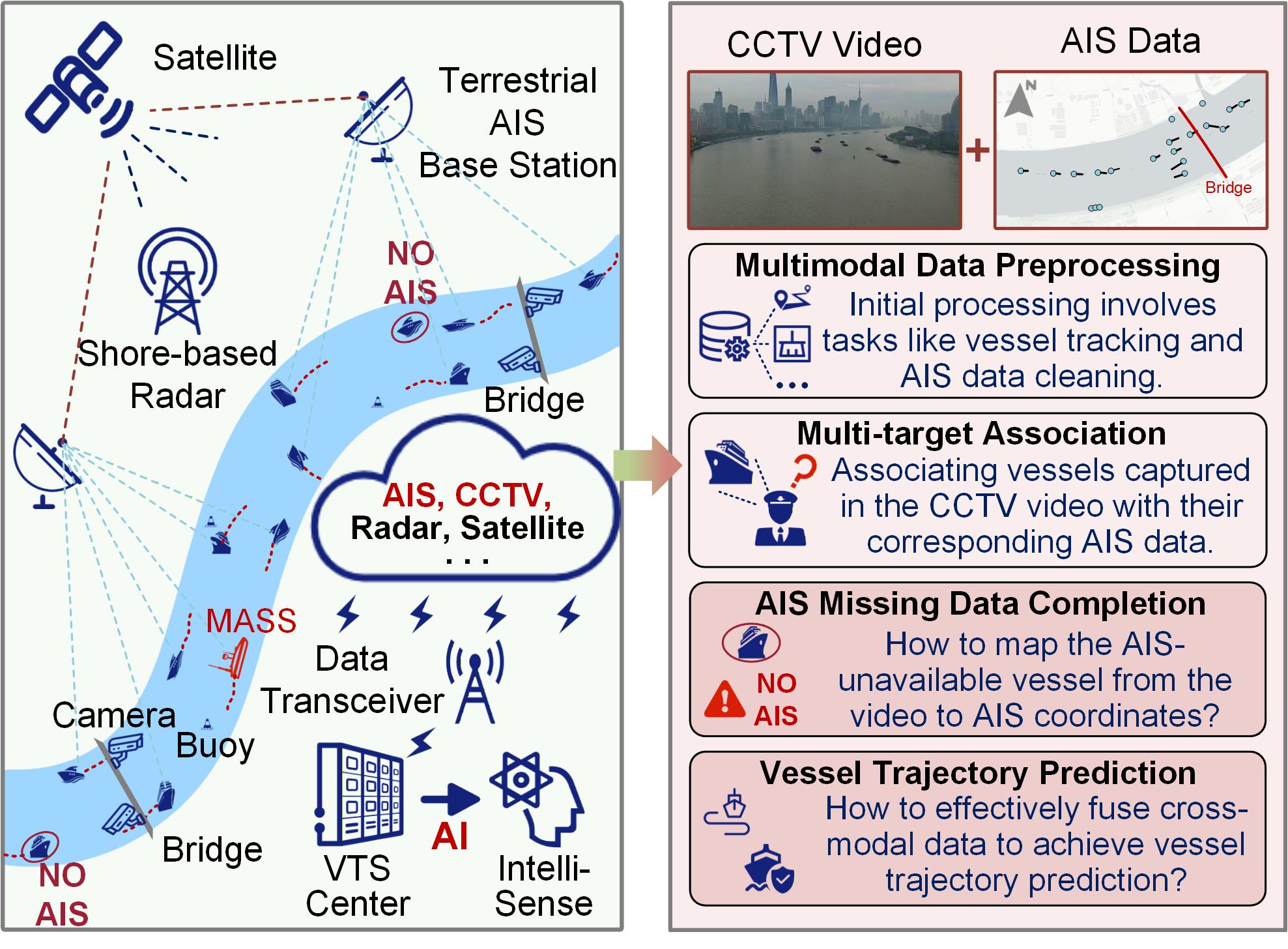}
        \caption{The MITS integrates advanced infrastructure and artificial intelligence-driven analytics to enable cross-modal interaction-based vessel trajectory prediction (e.g., using AIS and CCTV data), ultimately enhancing maritime intelligence and MASS operations for safe, sustainable navigation.}
        \label{Figure_Application}
    \end{figure}

    \IEEEPARstart{A}{chieving} omission-free and real-time situational awareness is a fundamental prerequisite for maritime intelligent transportation systems (MITS), serving as the basis for intelligent alarms and collision prevention to ensure navigational safety \cite{pietrzykowski2010maritime,zhao2026real}. As an essential component of MITS, maritime supervision involves monitoring and regulating vessel movements in areas such as ports or busy waterways \cite{li2023real}, leveraging technologies like the automatic identification system (AIS) \cite{liu2022stmgcn} and closed-circuit television (CCTV) \cite{liu2024real} to enforce safe navigation practices. Beyond ensuring compliance, MITS plays a critical role in advancing maritime autonomous surface ship (MASS) \cite{chang2021risk} and ship-shore collaboration \cite{chen2025cloud}. As illustrated in Fig. \ref{Figure_Application}, by synergizing multi-source data like AIS and CCTV to overcome individual sensor limitations, MITS not only guarantees operational safety and efficiency but also catalyzes the evolution of next-generation vessel traffic systems \cite{xiao2022next}.
    \begin{figure*}[t]
        \centering
        \setlength{\abovecaptionskip}{0.cm}
        \includegraphics[width=1.00\linewidth]{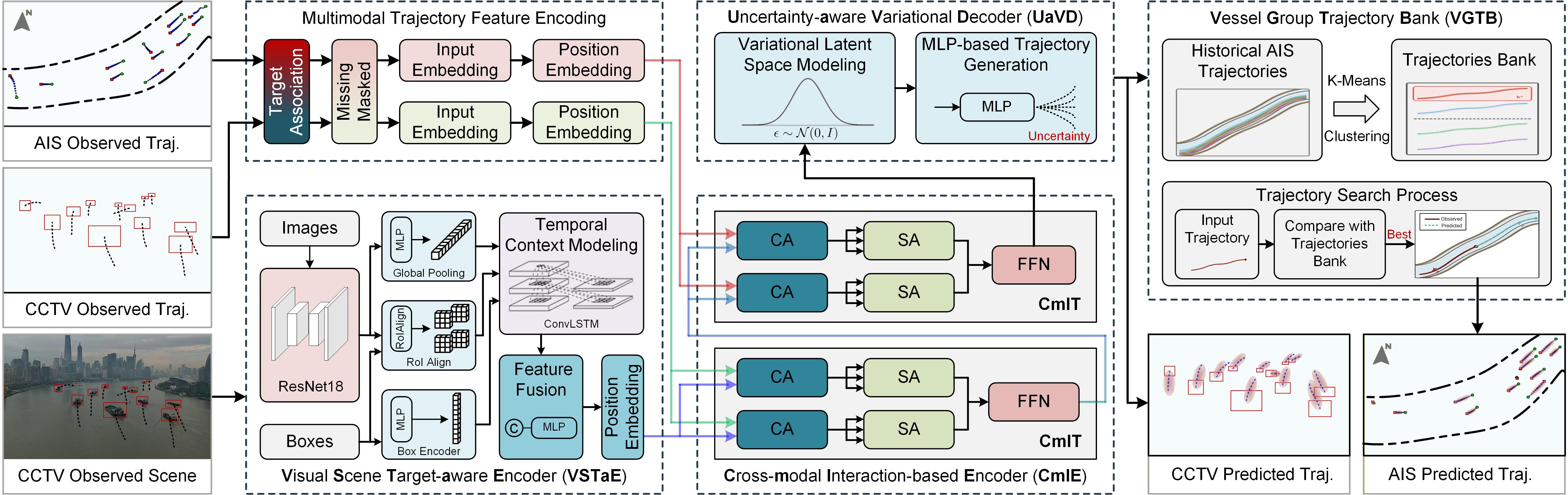}
        \caption{The flowchart of the proposed cross-modal interaction-based vessel trajectory prediction (named CmIVTP) framework. It consists of four main modules: the visual scene target-aware encoder (VSTaE) to extract environmental interaction features, the cross-modal interaction-based encoder (CmIE) to fuse AIS and CCTV data for modeling complex interactions, and the uncertainty-aware variational decoder (UaVD) to generate future trajectories. Additionally, a vessel group trajectory bank (VGTB) is constructed to improve the efficiency and accuracy of trajectory generation.}
        \label{figure:fig2}
    \end{figure*}

    In modern maritime surveillance, relying on a single data source is increasingly insufficient for comprehensive situational awareness \cite{lu2026graph}. Traditionally, AIS serves as the primary cooperative tracking tool, providing essential macroscopic vessel kinematics. However, as a radio-dependent system, its reliability is inherently vulnerable to signal loss, transmission delays, data sparsity, and the presence of uncooperative vessels \cite{tu2017exploiting,ribeiro2023ais}. CCTV acts as a non-cooperative visual sensor, capturing rich local semantics and environmental context independently of vessel broadcasts. In addition, high-performance graphics processing unit and rapid iterative optimization of learning methods make CCTV widely used in many fields of maritime affairs \cite{guo2023asynchronous}. Nevertheless, relying solely on real-time static and dynamic information from video footage is inadequate for ensuring safety supervision of the vessel when visual targets are occluded or suffer from imaging interference \cite{liu2024real}. To advance from passive monitoring to proactive collision avoidance, effective safety supervision fundamentally relies on multimodal data fusion to anticipate future vessel movements.
\subsection{Motivation}
    Multimodal-driven vessel trajectory prediction emerges as a critical prerequisite for MITS, which is essential for improving both the efficiency and safety of navigation in complex waterways \cite{zhang2022vessel}. AIS-based trajectory prediction methods \cite{li2023ais} have been used in different navigable waters. However, AIS limitations make it challenging to effectively monitor vessels navigating complex restricted waters, which include
    \begin{itemize}
        \item AIS may provide incomplete or unavailable data which limits the ability to track vessels accurately.
        \item AIS may experience delays in data transmission which hinders timely trajectory predictions.
        \item AIS may be intentionally turned off or not installed on a small number of vessels to evade detection.
    \end{itemize}

    Although previous efforts have explored AIS latency \cite{lu2026uncertainty,bao2026resilient}, tackling the complete unavailability of AIS signals remains a critical hurdle. To mitigate such inherent limitations, CCTV video serves as an indispensable modality by providing high-resolution visual semantics and environmental context. Currently, significant progress has been made in multimodal data-driven trajectory prediction for person \cite{meng2022forecasting,yang2023long} and vehicle \cite{chen2022intention,ren2024emsin}. However, transferring land-based methods to the maritime domain is sub-optimal. Unlike lane-constrained vehicles, vessels navigate in unstructured, open-water environments where their movements are governed by complex hydrodynamic behaviors and implicit interactions with the surroundings. For instance, while AIS provides sparse, macroscopic kinematic priors, it is challenging to capture the dense, microscopic visual cues necessary to perceive uncooperative vessels or complex spatial-temporal interactions in congested waterways \cite{bonci2019steering}. Consequently, trajectory prediction can be significantly enhanced by fusing macroscopic AIS kinematics with microscopic CCTV semantics to robustly capture complex vessel dynamics and compensate for the missing data of a single modality. In addition, the absence of synchronized AIS-CCTV datasets severely hinders the training of such models, making a foundational benchmark urgently needed.
\subsection{Contribution}
    \begin{figure*}[t]
        \centering
        \setlength{\abovecaptionskip}{0.cm}
        \includegraphics[width=1.00\linewidth]{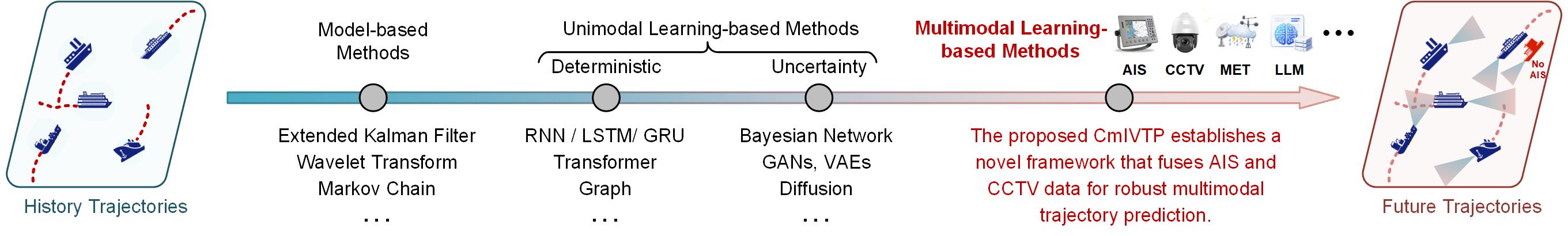}
        \caption{Vessel trajectory prediction has evolved from traditional model-based methods to unimodal, and subsequently to advanced multimodal learning methods. Advancing this trend, our CmIVTP proposes a novel multimodal framework fusing AIS and CCTV data, ensuring robust prediction through cross-modal compensation, even when single-modal data (e.g., AIS) is sparse or completely missing.}
        \label{figure:relatedwork}
    \end{figure*}
    To mitigate these challenges, as shown in Fig. \ref{figure:fig2}, we propose a \textit{cross-modal interaction-based} vessel trajectory prediction (named CmIVTP) framework that effectively fuses heterogeneous data from AIS and CCTV to model the complex interactions between vessel dynamics and environmental constraints. Specifically, we propose a visual scene target-aware encoder (VSTaE) to extract object-level scene features, effectively modeling vessel-environment interactions. Furthermore, we introduce the cross-modal interaction-based encoder (CmIE) that integrates AIS-derived motion features, CCTV-based environmental features, and scene representation features. We propose an uncertainty-aware variational decoder (UaVD) that shifts the prediction paradigm from a deterministic mapping to probabilistic distribution modeling. By learning a latent space to represent unobserved maneuvering intentions, it effectively captures multi-modal uncertainties to predict diverse future trajectories. To further enhance trajectory prediction, we construct a vessel group trajectory bank (VGTB) by clustering historical AIS trajectories into representative motion patterns, providing an efficient and scalable approach for candidate trajectory generation. Additionally, we introduce the maritime multimodal dataset plus (Maritime-MmD$^+$), a large-scale dataset that synchronizes AIS and CCTV video data, providing robust support for multimodal trajectory prediction. Extensive experimental results demonstrate that CmIVTP achieves better performance on vessel trajectory prediction benchmarks, validating its effectiveness and robustness in complex maritime scenarios. The main contributions of this work are summarized as follows
    \begin{itemize}
		\item  We propose a novel cross-modal interaction-based vessel trajectory prediction framework (named CmIVTP) that integrates AIS dynamics and CCTV context, addressing the challenge of multimodal data fusion for accurate and context-aware vessel trajectory modeling and prediction.
		\item We propose a VSTaE to capture vessel-environment interactions, a CmIE to integrate kinematic and visual features, a UaVD to model trajectory uncertainty via variational inference, and a VGTB to cluster AIS trajectories into representative patterns for motion prediction.
		\item The proposed CmIVTP demonstrates highly competitive performance in multimodal-driven vessel trajectory prediction. In addition, we construct the Maritime-MmD$^+$ to support research on multimodal data-driven vessel behavior analysis and maritime traffic pattern recognition.
    \end{itemize}
\subsection{Organization}
    The remainder of this work is organized as follows: The current studies on trajectory prediction methods are reviewed in Section \ref{relatedwork}. In Section \ref{sec:definition}, we initially define the research problem. Our method is detailedly described in Section \ref{sec:ourmethod}. Experimental results and discussion are provided in Section \ref{exper}. Section \ref{conc} concludes this work.
\section{Related Work}\label{relatedwork}
    This section provides a detailed review of vessel trajectory prediction methods across three categories: model-, unimodal learning-, and multimodal learning-based methods, while explicitly highlighting the differences from prior research.
\subsection{Model-based Methods}
    Model-based methods primarily rely on vessel motion rules and physical factors (e.g., size, draft, and inertia) to establish mathematical models describing vessel operations \cite{zhou2019review}, including curve \cite{perera2010ocean} and vessel models \cite{semerdjiev2000variable}. For example, Perera \textit{et al}. \cite{perera2010ocean} utilized an extended Kalman filter combining curvilinear and linear models to predict vessel kinematics. However, Kalman filtering and linear regression often suffer from large time delays and limited accuracy. To address complex dynamics, Zhang \textit{et al}. \cite{zhang2019wavelet} combined wavelet transforms with hidden Markov models to estimate future positions from trajectory sequences. Similarly, Guo \textit{et al.} \cite{guo2018trajectory} incorporated environmental parameters (e.g., tides, weather) into a k-order multivariate Markov chain for trajectory prediction. Despite these efforts, model-based methods primarily focus on the isolated motion of a single target vessel. They are generally challenging to account for multi-vessel interactions, which significantly limits their applicability in complex, real-world navigation scenarios \cite{huang2020ship}.
\subsection{Unimodal Learning-based Methods}
    Benefiting from the robust nonlinear fitting and parallel computing capabilities of deep learning, unimodal learning-based methods are primarily divided into deterministic and uncertainty methods \cite{li2023ais}.
\subsubsection{Deterministic Learning}
    Deterministic methods generate a single predicted path based on historical trajectories and environmental data. Recurrent neural networks (RNN) \cite{capobianco2021deep,jurkus2023application} and variants like long short-term memory (LSTM) \cite{liu2022deep,chondrodima2023efficient} and gated recurrent unit (GRU) \cite{chen2022fb} are widely adopted in vessel trajectory prediction due to their excellent time-series learning capabilities. To optimize complex hyperparameters and improve accuracy, clustering and genetic algorithms are often employed. For instance, Murray \textit{et al.} \cite{murray2021ais} pre-processed historical vessel behaviors into geographical clusters. To capture multi-vessel interactions, researchers have modeled individual trajectory interactions using social pooling modules \cite{liu2022deep}. Additionally, Capobianco \textit{et al.} \cite{capobianco2021deep} introduced an attention mechanism to learn motion state relationships while preserving input spatiotemporal structures. Furthermore, graph networks excel at modeling complex vessel interactions by handling non-Euclidean structured data. Specifically, Wang \textit{et al.} \cite{wang2024vessel} utilized a spatiotemporal graph convolutional network to extract motion features, while Yang \textit{et al.} \cite{yang2025enhancing} proposed a graph-based model to capture interactions, thereby concurrently improving trajectory prediction and risk perception.
\subsubsection{Uncertainty Learning}
    Uncertainty trajectory prediction estimates the probabilistic distribution of potential paths by modeling the stochastic characteristics of maritime dynamics. To capture time-varying dynamics, Jia \textit{et al.} \cite{jia2024multiple} combined variational Kalman filters with GRU networks within a Bayesian framework. Gaussian models \cite{rong2019ship,sorensen2022probabilistic} are frequently coupled with deep networks to learn positional uncertainties. For example, Sørensen \textit{et al.} \cite{sorensen2022probabilistic} utilized an eleven-dimensional Gaussian distribution alongside a mixture density network to predict future positions. Similarly, Guo \textit{et al.} \cite{guo2023toward} proposed Gaussian-sampled latent vectors with adversarial learning for uncertain trajectory prediction. Generative models are also widely adopted to simulate diverse movement behaviors. Chen \textit{et al.} \cite{chen2024regional} applied generative adversarial networks (GANs) with pooling modules to aggregate social interactions among neighboring vessels and generate diverse trajectories. Meanwhile, Han \textit{et al.} \cite{han2023interaction} utilized a conditional variational autoencoder (CVAE) framework to learn marine vessel movements. Furthermore, Han \textit{et al.} \cite{han2025probabilistic} integrated attention mechanism with diffusion models to address trajectory uncertainty via an inverse motion diffusion process.
\subsection{Multimodal Learning-based Methods}
    While AIS kinematics traditionally dominate prediction pipelines, vessel motion is heavily shaped by diverse exogenous environmental factors. Consequently, integrating multiple data sources has become crucial for accurate forecasting. TripleConvTransformer \cite{huang2022tripleconvtransformer} took an early step by combining AIS-derived kinematic sequences with discretized meteorological (MET) fields using a multi-branch convolutional-attentional architecture. Each branch extracts modality-specific features to capture complex interactions between vessel dynamics and external conditions. Building upon this foundation, Xiao \textit{et al.} \cite{xiao2024adaptive} broadened multimodal research by fusing satellite AIS data with environmental variables through dedicated extraction networks and an adaptive fusion block. Subsequent research has increasingly shifted toward systematic and fine-grained multimodal fusion strategies. For instance, MDL-TP \cite{luo2025multimodal} establishes a unified multimodal learning pipeline by jointly constructing hybrid datasets and employing end-to-end fusion architectures. Complementarily, ST-FEiTNet \cite{zhou2025st} focuses on the joint representation of motion and environment under complex sea states. It dynamically integrates sea-state information with vessel motion features through cross-modal frequency enhancement. Bao \textit{et al.} \cite{bao2026resilient} proposed a graph-driven trajectory prediction framework fusing CCTV and AIS data, which utilizes visual-aided data imputation and cross-modal graph attention to effectively resolve the missing data and interference issues of single-source modalities. Transcending traditional physical constraints, contemporary approaches leverage large language models (LLMs) to seamlessly fuse high-level semantic modalities with low-level numerical trajectories. Specifically, models including SEMINT \cite{chen2025semint} and LLM4STP \cite{jiao2026llm4stp} utilize cross-modal reasoning mechanisms to assimilate intention-aware contexts and multidimensional features. Consequently, foundational architectures like AIS-LLM \cite{park2025ais} validate the potential of LLMs in processing and aligning heterogeneous multimodal inputs for enhanced maritime situational awareness.
\subsection{Difference from Previous Works}
    \begin{figure}[t]
        \centering
        \setlength{\abovecaptionskip}{0.cm}
        \includegraphics[width=1.00\linewidth]{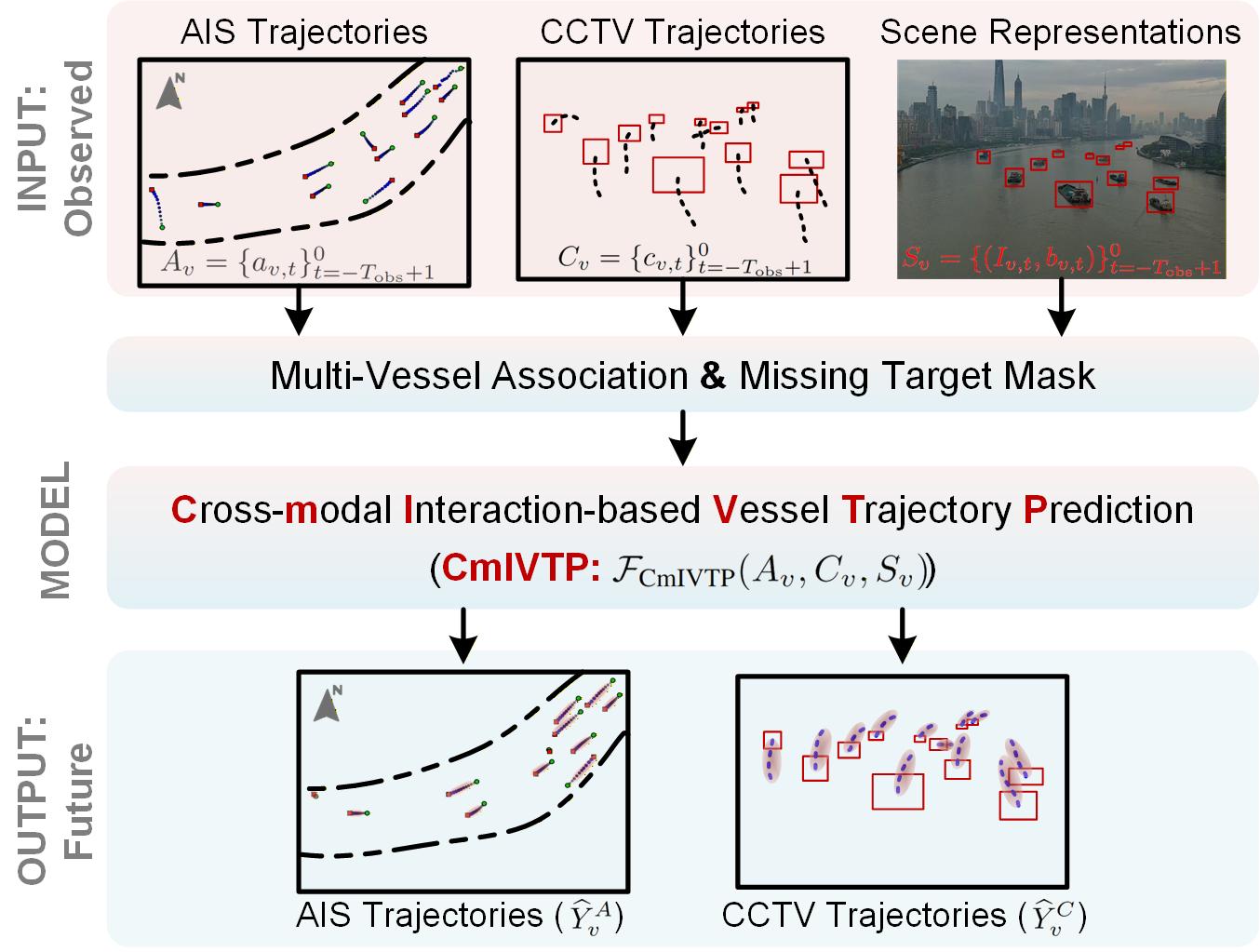}
        \caption{Problem setup for cross-modal interaction-based vessel trajectory prediction. Given AIS trajectories, CCTV trajectories, and scene representations, the goal is to predict future vessel trajectories using cross-modal interactions to address modality-specific dynamics and sensor limitations.}
        \label{Figure_problem}
    \end{figure}
    As shown in Fig. \ref{figure:relatedwork}, previous prediction methods primarily rely on single-source AIS data, which suffers from inherent limitations such as low update frequency, potential data loss, and inability to capture fine-grained maneuvering behaviors in complex maritime scenarios \cite{guo2023asynchronous,lu2026graph}. Unlike road vehicles constrained by lanes \cite{mo2024heterogeneous}, vessels may perform maritime-specific behaviors such as complex evasive maneuvers and continuous course alterations in curved waterways \cite{bruns2023learning}. Such behavioral complexity, combined with the sparse and delayed nature of AIS data, necessitates the integration of complementary CCTV information sources. Distinct from existing unimodal paradigms, our framework overcomes these limitations by integrating multimodal information from AIS data and CCTV video, enabling environmentally consistent and uncertainty-aware modeling of vessel behaviors.
\section{Initial Definition}\label{sec:definition}
   The inherent limitations of individual data modalities pose significant challenges to robust trajectory prediction. Integrating complementary sources is thus essential, enabling synergistic capture of global kinematic patterns and fine-grained local interactions for reliable predictions in complex waterways. In our prior work \cite{guo2023asynchronous,lu2026uncertainty,lu2026graph}, we have systematically addressed data preprocessing and spatio-temporal alignment, along with multi-target association and missing target masks in AIS and CCTV data. Therefore, this work focuses on \textit{the cross-modal interaction-based vessel trajectory prediction} task. As shown in Fig. \ref{Figure_problem}, we consider a target vessel $v$ observed over $T_{\mathrm{obs}}$ time steps, defined as follows
    \begin{itemize}
		\item AIS Trajectories: $A_v = \{ a_{v,t} \}_{t=-T_{\mathrm{obs}}+1}^{0}$, with $a_{v,t} = (\mathrm{lon}_{v,t}, \mathrm{lat}_{v,t}) \in \mathbb{R}^2$ denotes the vessel's geographic position at time $t$, consisting of longitude and latitude.
		\item CCTV Trajectories: $C_v = \{ c_{v,t} \}_{t=-T_{\mathrm{obs}}+1}^{0}$, with $c_{v,t} = (x^{\mathrm{mid}}_{v,t}, y^{\mathrm{mid}}_{v,t}) \in \mathbb{R}^{2}$ denotes the center point of the vessel's bounding box in the video frame at time $t$.
		\item Scene Representations: $S_v = \{ (I_{v,t}, b_{v,t}) \}_{t=-T_{\mathrm{obs}}+1}^{0}$, with $I_{v,t} \in \mathbb{R}^{3 \times H \times W}$ is the global semantic feature, and $b_{v,t} = [x^{\min}_{v,t}, y^{\min}_{v,t}, x^{\max}_{v,t}, y^{\max}_{v,t}] \in \mathbb{R}^4$ denotes the top-left and bottom-right corners of vessel's bounding box in the video frame at time $t$.
    \end{itemize}

    The final goal of this work is to predict the vessel's uncertain future trajectories over $T_{\mathrm{fut}}$ timesteps conditioned on the multimodal history, with modality-specific future targets and predictions, which can be given as
    \begin{equation}
        \widehat{Y}_v^{A}, \widehat{Y}_v^{C} = \mathcal{F}_{\boldsymbol{\text{CmIVTP}}}(A_v, C_v, S_v),
    \end{equation}
    where $\mathcal{F}_{\text{CmIVTP}}(\cdot)$ is our proposed cross-modal interaction-based vessel trajectory prediction model, and $\widehat{Y}_v^{A}, \widehat{Y}_v^{C}$ are the predicted AIS and CCTV trajectory sequences, respectively.
    \begin{algorithm}[t]
    \caption{The Proposed CmIVTP Framework}
    \label{alg:CmIVTP}
        \begin{algorithmic}[1]
            \Require 
            \State $A_v$: AIS trajectory over observation period
            \State $C_v$: CCTV trajectory over observation period
            \State $S_v$: Scene representations over observation period
            \State $\mathcal{Z}_{\mathrm{bank}}$: Optional vessel trajectory bank for refinement
            \State $K \in \mathbb{N}$: Number of modes for uncertainty modeling
            \State $T_{\mathrm{fut}} \in \mathbb{N}$: Prediction horizon (number of future timesteps)  
            \Ensure
            \State $\widehat{Y}_v^{A} = \{\widehat{a}_{v,t}\}_{t=1}^{T_{\mathrm{fut}}}$: Predicted AIS trajectory
            \State $\widehat{Y}_v^{C} = \{\widehat{c}_{v,t}\}_{t=1}^{T_{\mathrm{fut}}}$: Predicted CCTV trajectory
    
            \Statex \texttt{// Step 1: Visual Scene Target-aware Encoding}
            \For{$t = -T_{\mathrm{obs}}+1$ to $0$} 
                \State Extract spatial features: 
                
                $f_{v,t}^{\text{roi}} \leftarrow \text{Fusion}(f^{\text{tar}}, f^{\text{glo}}, f^{\text{bbox}})$
                \State Update temporal context: $\tilde{f}_{v,t}^{\text{temp}} \leftarrow \text{ConvLSTM}(F_{v,t})$
                \State Apply temporal weighting: $f_{v,t}^{\text{temp}} \leftarrow w_t \cdot \tilde{f}_{v,t}^{\text{temp}}$
            \EndFor
            \State Fuse spatiotemporal features: 
            
            $F_{v}^{\text{VSTaE}} \leftarrow \text{MLP}([F_v^{\text{roi}} \oplus F_v^{\text{temp}}])$
    
            \Statex \texttt{// Step 2: Cross-modal Interaction-based Encoding}
            \State Encode AIS trajectory features:
            $F_{v}^{A} = g_{\text{AIS}}(A_v)$
            \State Encode CCTV trajectory features:
            $F_{v}^{C} = g_{\text{CCTV}}(C_v)$
            \State Fuse multi-modal features via cascaded CMIT:
            
            $F_{v}^{\text{fus}} = \text{CMIT}(F_{v}^{A}, F_{v}^{C}, F_{v}^{\text{VSTaE}})$
    
            \Statex \texttt{// Step 3: Uncertainty-aware Variational Decoding}
            \For{$k = 1$ to $K$} \Comment{Iterate over prediction modes}
                \State Sample stochastic latent variable: 
                
                $z_k \sim \mathcal{N}(\mu, \operatorname{diag}(\sigma^2))$
                \State Predict base AIS and CCTV trajectories via MLP:
                $\widehat{Y}_v^{A\_base,k}, \widehat{Y}_v^{C,k} = \text{MLP}(F_{v}^{\text{fus}}, z_k)$
            \EndFor

            \Statex \texttt{// Step 4: Vessel Group Trajectory Bank Refinement}
            \If{\(\mathcal{B}_{\text{ais}} \neq \emptyset\)} 
            \State Retrieve prior: \(\widetilde{Y}_{k^\ast} \leftarrow \text{Search}(\mathcal{B}_{\text{ais}}, A_v)\)
            \State Refine predictions:
            
            \(\widehat{Y}_v^{(A),k} = \text{AdaptiveFusion}(\widehat{Y}_{v,\text{base}}^{(A),k}, \widetilde{Y}_{k^\ast}, F_{v}^{\text{fus}})\)
            \Else
            \State \(\widehat{Y}_v^{(A),k} = \widehat{Y}_{v,\text{base}}^{(A),k}\)
        \EndIf
        
        \State \Return \(\widehat{Y}_v^{(A)} = \{\widehat{Y}_v^{(A),k}\}_{k=1}^{K}\), \(\widehat{Y}_v^{C} = \{\widehat{Y}_v^{C,k}\}_{k=1}^{K}\)
    
        \end{algorithmic}\label{alg1}
    \end{algorithm}
\section{Our Method}\label{sec:ourmethod}
\subsection{Overview of CmIVTP}
    This work proposes a novel CmIVTP framework that combines AIS trajectories data, CCTV trajectories data, and CCTV scene representations features to overcome single-source maritime limitations. As shown in Fig. \ref{figure:fig2} and Alg. \ref{alg1}, the framework first utilizes the VSTaE to extract object- and pixel-level scene representations, providing critical spatial and contextual details for precise vessel behavior understanding. The CmIE then employs cross-modal attention to deeply interact and fuse multi-modal features, effectively compensating for modality-specific data loss (e.g., missing AIS). Subsequently, the UaVD models unobserved maneuvering intentions to encode trajectory uncertainty. Finally, the VGTB retrieves similar historical trajectories, refining predictions through learnable offsets, thereby improving accuracy and robustness.
\subsection{Visual Scene Target-aware Encoder}
    \begin{figure}[t]
        \centering
        \setlength{\abovecaptionskip}{0.cm}
        \includegraphics[width=1.00\linewidth]{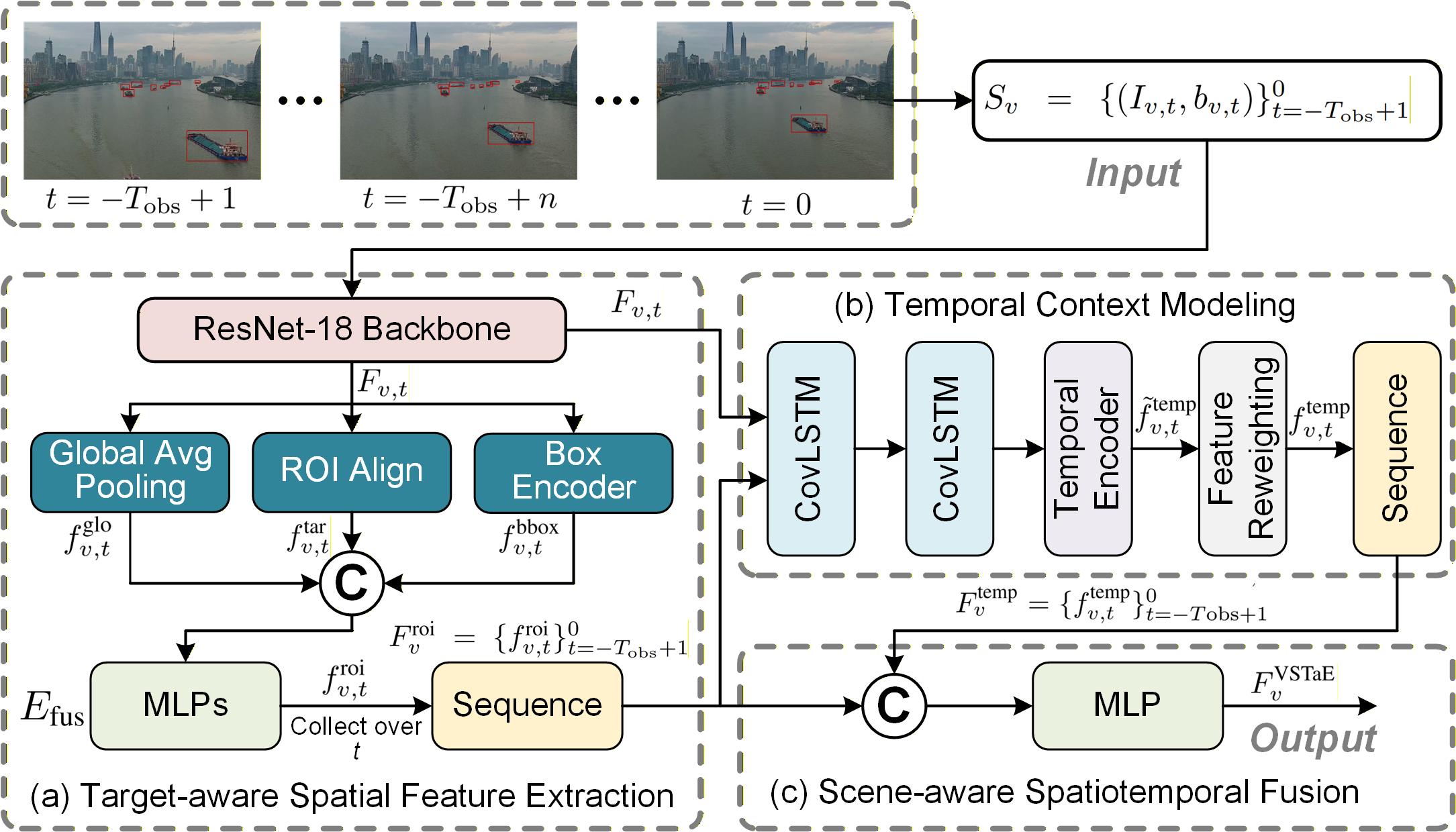}
        \caption{The pipeline of the VSTaE. It extracts features from image sequences using target-aware spatial feature extraction, captures dynamics via temporal context modeling with ConvLSTM, and combines them through scene-aware spatiotemporal fusion to output representations for trajectory prediction.}
        \label{Figure_vstae}
    \end{figure}
    To adapt standard visual encoding for the specific demands of maritime environments, as shown in Fig. \ref{Figure_vstae}, we propose a VSTaE to compute spatiotemporal representations $F_{v,\text{VSTaE}} \in \mathbb{R}^{T_{\mathrm{obs}} \times d}$ for the target vessel $v$ over the observation period $T_{\mathrm{obs}}$, using scene representations $S_v = \{(I_{v,t}, b_{v,t})\}_{t=-T_{\mathrm{obs}}+1}^{0}$. By extracting spatial and temporal features, the VSTaE ensures that vessel-environment interactions, such as proximity to other vessels, navigational hazards, and environmental constraints, are effectively captured. The VSTaE integrates three core components: the target-aware spatial feature extraction for vessel-specific and scene-level spatial features, the temporal context modeling for spatiotemporal dependencies, and the scene-aware spatiotemporal fusion for combining spatial and temporal features into a unified representation for trajectory predictions.
\subsubsection{Target-aware Spatial Feature Extraction}
    For each timestep $t$, the target-aware spatial feature extraction processes the input frame $I_{v,t}$ through a ResNet-18 backbone \cite{he2016deep} to generate global feature maps $F_{v,t}$. To capture maritime-specific vessel-environment constraints rather than generic visual semantics, the module extracts and fuses three types of features for the bounding box $b_{v,t}$ of the target vessel, which can be given as
    \begin{equation}
        f_{v,t}^{\text{roi}} = E_{\text{fus}}([f_{v,t}^{\text{tar}} \oplus f_{v,t}^{\text{glo}} \oplus f_{v,t}^{\text{bbox}}]),
    \end{equation}
    where $f_{v,t}^{\text{tar}} \in \mathbb{R}^d$ is the target-specific embedding extracted via RoI Align with output size $7 \times 7$ and spatial scale $\frac{1}{32}$, $f_{v,t}^{\text{glo}} \in \mathbb{R}^{d/2}$ is the global context obtained from global average pooling over $F_{v,t}$, and $f_{v,t}^{\text{bbox}} \in \mathbb{R}^{64}$ is the bounding box positional encoding processed through MLP layers. The fusion module $E_{\text{fus}}$ is a two-layer MLP that aggregates these features, effectively combining local semantics, global context, and spatial positional information.
\subsubsection{Temporal Context Modeling}
    The sequence of global feature maps $\{F_{v,t}\}_{t=-T_{\mathrm{obs}}+1}^{0}$ is then organized in temporal-first format and processed by the temporal context modeling module, which is implemented as a two-layer Convolutional long short-term memory (ConvLSTM) \cite{shi2015convolutional}. The temporal context modeling module models the temporal evolution of vessel dynamics and scene changes over the observation period, capturing spatiotemporal dependencies critical for trajectory prediction. The outputs of the final ConvLSTM layer are encoded by $E_{\text{temp}}(\cdot)$ to produce compact temporal features for each timestep, which can be given as
    \begin{equation}
        \tilde{f}_{v,t}^{\text{temp}} = E_{\text{temp}}(\text{ConvLSTM}(F_{v,t})).
    \end{equation}

    To emphasize recent observations while retaining the long-term context, exponential temporal weighting $w_t$ is applied to the temporal features, i.e.,
    \begin{equation}\label{eq:ftemp}
        f_{v,t}^{\text{temp}} = w_t \cdot \tilde{f}_{v,t}^{\text{temp}}, \quad
        w_t = \exp(\phi \cdot t),
    \end{equation}
    where $\phi =0.1$ is a weighting factor controlling the rate of decay. The temporal weighting mechanism prioritizes short-term dynamics to capture sudden maneuvers, while preserving long-term trends to account for the high inertia typical of maritime vessels.
    \begin{figure}[t]
        \centering
        \setlength{\abovecaptionskip}{0.cm}
        \includegraphics[width=1.00\linewidth]{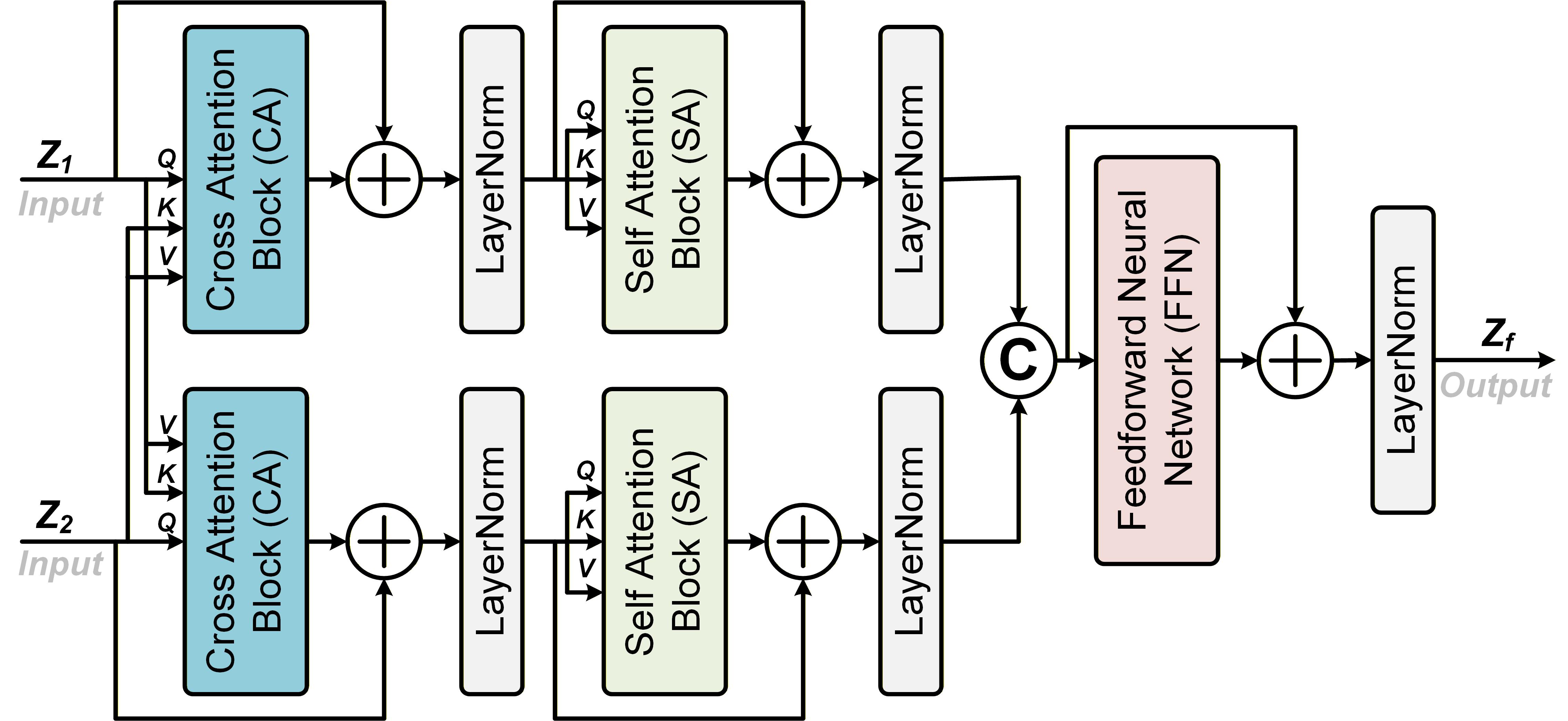}
        \caption{The pipeline of the cross-modal interaction Transformer (CMIT). It comprises cross-attention (CA), self-attention (SA), and feed-forward network (FFN) modules, which enables the model to simultaneously capture intra-modal semantics and inter-modal interactions.}
        \label{Figure_cmie}
    \end{figure}
\subsubsection{Scene-aware Spatiotemporal Fusion}
    Finally, the spatial features $F_v^{\text{roi}} = \{f_{v,t}^{\text{roi}} \}_{t=-T_{\mathrm{obs}}+1}^{0}$ and temporal features $F_v^{\text{temp}} = \{ f_{v,t}^{\text{temp}} \}_{t=-T{\mathrm{obs}}+1}^{0}$ are fused via an MLP to produce the integrated spatiotemporal representation, i.e.,
    \begin{equation}
        F_{v}^\text{VSTaE} = \text{MLP}([F_v^{\text{roi}} \oplus F_v^{\text{temp}}]),
    \end{equation}
    where $\oplus$ denotes feature concatenation. The integrated representation effectively combines vessel-specific features, global scene context, and temporal evolution, enabling the framework to generate trajectory predictions that are both dynamically consistent and environmentally feasible.
\subsection{Cross-modal Interaction-based Encoder}
    To explicitly align vessel dynamics with visual environmental constraints rather than performing a naive feature concatenation, the CmIE is constituted by the cross-modal interaction Transformer (CMIT), which enables the model to sequentially capture intra-modal semantics and inter-modal interactions. As shown in Fig. \ref{Figure_cmie}, the module comprises self-attention (SA), cross-attention (CA), and feed-forward network (FFN) modules. Let $Z_1, Z_2 \in \mathbb{R}^{T \times d}$ represent two input feature sets over $T$ timesteps, where $Z_1$ acts as the primary query and $Z_2$ serves as the memory. Unlike standard symmetric fusion approaches, the CMIT is designed as an asymmetric architecture to inject contextual memory into a primary trajectory representation.
\subsubsection{Interaction and Fusion Process}
    The processing pipeline consists of three sequential stages: intra-modal refinement, cross-modal interaction, and feature transformation. First, the primary modality refines temporal dependencies via self-attention for trajectory smoothness. Second, it queries the memory modality to incorporate interactive dynamics, bounding the vessel's motion with scene-level navigational constraints. Formally, given the primary feature $Z_1$ and memory feature $Z_2$, the intermediate features are updated as
    \begin{equation}
        \begin{aligned}
            \bar{Z}_1 &= \text{LN}(Z_1 + \text{SA}(Z_1, Z_1)), \\
            \tilde{Z}_1 &= \text{LN}(\bar{Z}_1 + \text{CA}(\bar{Z}_1, Z_2)),
        \end{aligned}
    \end{equation}
    where $\text{LN}(\cdot)$ denotes Layer Normalization. Finally, the fused feature $Z_f$ is refined through a feed-forward network, i.e.,
    \begin{equation}
        Z_f = \text{LN}(\tilde{Z}_1 + \text{FFN}(\tilde{Z}_1)).
    \end{equation}
\subsubsection{Attention Mechanism}
    Both CA and SA are implemented using scaled dot-product attention, i.e.,
    \begin{equation}
        \text{Attn}(Q, K, V) = \text{softmax}\left(\frac{Q K^\top}{\sqrt{d_k}}\right)V,
    \end{equation}
    where $Q$, $K$, and $V$ denote the query, key, and value matrices, respectively, and $d_k$ is the dimension of the keys. Specifically, in SA, the primary feature $Z_1$ provides $Q, K, V$ to refine internal temporal dependencies. In CA, the updated $\bar{Z}_1$ acts as the query $Q$ to dynamically retrieve relevant contexts from the memory $Z_2$ (serving as $K$ and $V$). Ultimately, this asymmetric fusion yields a unified representation where vessel dynamics and visual contexts are deeply intertwined.
\subsection{Uncertainty-aware Variational Decoder}
    \begin{figure}[t]
        \centering
        \setlength{\abovecaptionskip}{0.cm}
        \includegraphics[width=1.00\linewidth]{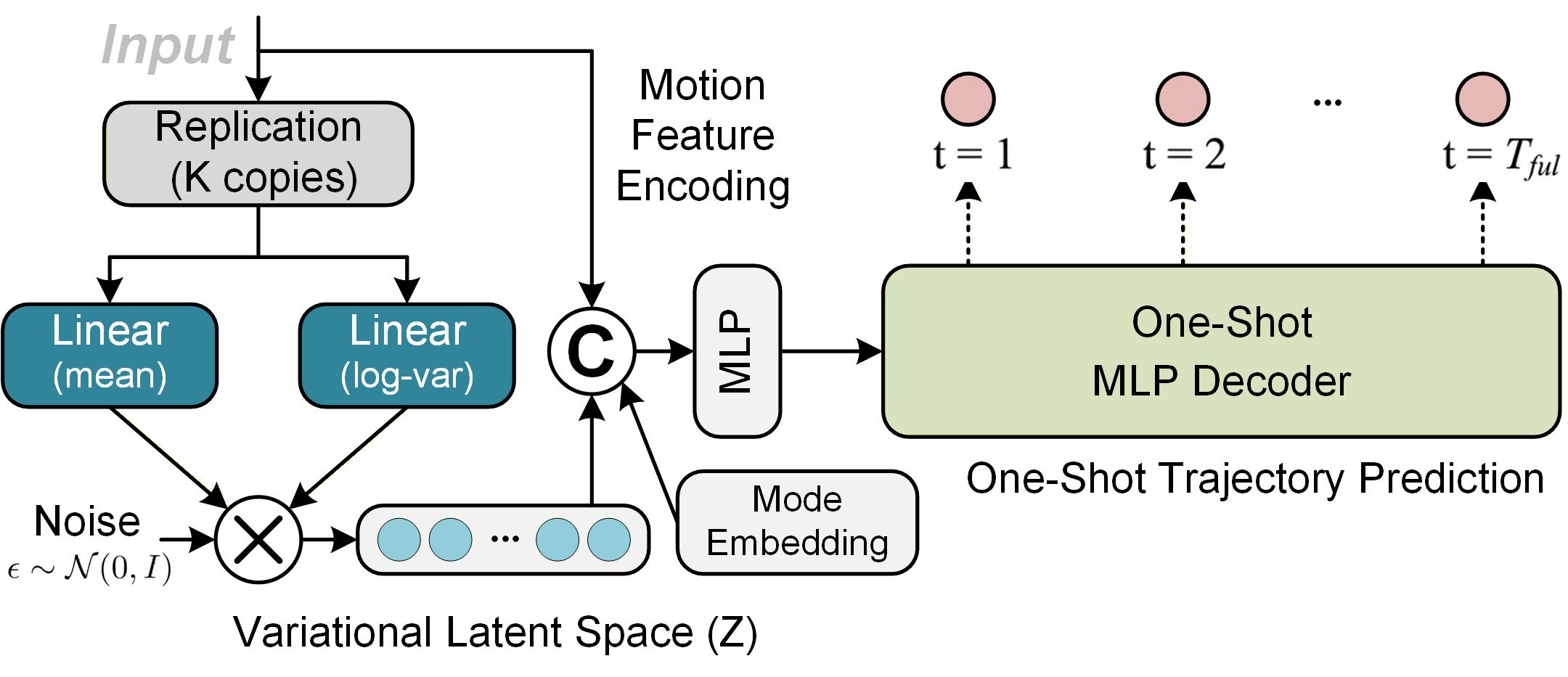}
        \caption{The pipeline of the uncertainty-aware variational decoder (UaVD). It samples multimodal latent variables from the variational latent space, decodes them using a MLP to iteratively generate future trajectories, and outputs multimodal trajectory predictions.}
        \label{Figure_vgtb}
    \end{figure}
    Traditional trajectory prediction methods often formulate the task as a deterministic mapping $Y = f(X)$, which struggles to account for the inherent stochasticity of vessel movements. To address this, we transition to a probabilistic distribution modeling problem $P(Y|X, z)$ by proposing a UaVD. Drawing from the conditional variational autoencoder framework, the UaVD introduces a latent space to model unobserved factors, such as human intentions or sudden environmental disturbances. The latent variable $z$ acts as a mathematical proxy for these unknown variables. By sampling different $z$ values from the learned latent distribution, the model captures "multi-modal" uncertainties. Specifically, each distinct sample of $z$ decodes into a different, physically plausible future trajectory mode, such as maintaining course, turning, or slowing down. As shown in Fig. \ref{Figure_vgtb}, this enables the generation of spatiotemporal representations $F_{\text{UaVD}} \in \mathbb{R}^{K \times T_{\text{fut}} \times d}$ (where $K$ denotes the number of modes, $T_{\text{fut}}$ the prediction horizon, and $d$ the feature dimension), facilitating diverse and uncertainty-aware trajectory predictions. The network consists of two key components: a variational latent space modeling (VLSM) module to sample multimodal latent representations $Z \in \mathbb{R}^{K \times d}$, and an MLP-based decoder to generate future trajectories conditioned on sampled latent variables.
\subsubsection{Variational Latent Space Modeling}
    The latent space modeling integrates the features $F_{\text{enc}} \in \mathbb{R}^{d}$ from the encoder and a learnable mode embedding $E_k$ with a reparameterization trick to sample the latent variable $z_k \in \mathbb{R}^d$ for each mode $k \in \{1, \dots, K\}$, which can be given as
    \begin{equation}
        z_k = \mu(F_{\text{enc}}, E_k) + \epsilon \odot \sigma(F_{\text{enc}}, E_k), \quad \epsilon \sim \mathcal{N}(0, I),
    \end{equation}
    where $\mu(\cdot)$ and $\sigma(\cdot)$ are linear transformations parameterizing the mean and standard deviation of the variational distribution, and $I \in \mathbb{R}^{d \times d}$ is the identity matrix.
    \begin{algorithm}[t]
        \caption{Vessel Group Trajectory Bank (VGTB)}
        \label{alg:vgtb_opt}
        \begin{algorithmic}[1]
            \Require $\mathcal{A}$: Historical AIS; $A_v$: Target observation; $K$: Clusters; $\alpha, \beta$: Weights
            \Ensure $\widehat{Y}_v^{(A)}$: Predicted future AIS trajectory
    
            \Statex \texttt{// Phase 1: Offline Trajectory Bank Construction}
            \State Extract observed $\{X_i\}$ and future $\{Y_i\}$ segments from $\mathcal{A}$
            \State Extract normalized features $\{f_i\}$ from $\{X_i\}$ via shift and scale
            \State Cluster $\{f_i\}$ into $K$ groups $\{C_k\}_{k=1}^K$ using K-means
            \For{$k = 1$ to $K$}
                \State Find medoid: $i^* = \arg\min_{i \in C_k} \| f_i - \frac{1}{|C_k|} \sum_{j \in C_k} f_j \|_2$
                \State Add to bank $\mathcal{B}_{\text{ais}}$: $(\widetilde{X}_k, \widetilde{Y}_k, f_{i^*}) \leftarrow (X_{i^*}, Y_{i^*}, f_{i^*})$
            \EndFor
    
            \Statex \texttt{// Phase 2: Online Search, Refinement, and Fusion}
            \State Extract normalized feature $f_v$ for target $A_v$
            \State Find best match: $k^\ast = \arg\max_{k} \frac{f_v^\top f_k}{\|f_v\| \|f_k\| + \epsilon}$
            \State Retrieve prior prediction: $\widehat{Y}_{v,\text{prior}}^{(A)} = \widetilde{Y}_{k^\ast}$
            \State Generate base prediction $\widehat{Y}_{v,\text{base}}^{(A)}$ and offset $O_v$
            \State Refine prior: $\widehat{Y}_{v,\text{ref}}^{(A)} = \widehat{Y}_{v,\text{prior}}^{(A)} + \alpha O_v$
            \State Final fusion: $\widehat{Y}_v^{(A)} = \beta \widehat{Y}_{v,\text{base}}^{(A)} + (1 - \beta) \widehat{Y}_{v,\text{ref}}^{(A)}$
            
            \State \Return $\widehat{Y}_v^{(A)}$
        \end{algorithmic}
    \end{algorithm}
\subsubsection{MLP-based Trajectory Generation}
    Instead of step-by-step recurrent generation, the decoder generates future trajectories $F_{\text{pred}} \in \mathbb{R}^{K \times T_{\mathrm{fut}} \times d}$ efficiently by employing a MLP to decode the entire sequence simultaneously. For each mode $k$, the input feature $F_{\text{in}, k}$ is formed by concatenating $F_{\text{enc}}$, $z_k$, and $E_k$, which is then mapped to the future sequence, i.e.,
    \begin{equation}
        F_{\text{pred}, k} = \mathcal{R}(\text{MLP}(F_{\text{in}, k})),
    \end{equation}
     where $\text{MLP}(\cdot)$ expands the concatenated feature to a flat vector of dimension $T_{\mathrm{fut}} \cdot d$, and $\mathcal{R}(\cdot)$ is a reshaping operator that formats it into the temporal sequence of size $T_{\mathrm{fut}} \times d$. The final multimodal trajectory predictions $F_{\text{UaVD}}$ are obtained by stacking the outputs for all $K$ modes. This latent conditioning promotes structured multimodality, enhancing robustness in uncertainty-aware forecasting.
\subsection{Vessel Group Trajectory Bank Module}
    The vessel group trajectory bank (VGTB) module constructs representative group trajectories from AIS data to support prediction of the future AIS trajectory $\widehat{Y}_v^{(A)}$ for the target vessel $v$. Its core functions include bank initialization, trajectory search, and trajectory update, using the AIS trajectory dataset $\mathcal{A} = \{ A_i \}_{i=1}^N$, where each $A_i$ contains longitude and latitude coordinates over a sequence of timesteps.
\subsubsection{Trajectory Bank Initialization}
    Each trajectory $A_i$ in $\mathcal{A}$ is divided into an observed trajectory $X_i = \{ x_{i,t} \}_{t=-T_{\mathrm{obs}}+1}^{0}$, $x_{i,t} \in \mathbb{R}^2$, containing longitude and latitude coordinates, and a future trajectory $Y_i = \{ y_{i,t} \}_{t=1}^{T_{\mathrm{fut}}}$, $y_{i,t} \in \mathbb{R}^2$, such that
    \begin{equation}
        A_i = \{ X_i; Y_i \}, \quad X_i \in \mathbb{R}^{T_{\mathrm{obs}} \times 2}, \quad Y_i \in \mathbb{R}^{T_{\mathrm{fut}} \times 2}.
    \end{equation}

    This yields two sets,that is, the observed set $\mathcal{X} = \{ X_i \}_{i=1}^N$ and the future set $\mathcal{Y} = \{ Y_i \}_{i=1}^N$.
\subsubsection{Constructing the Trajectory Bank}
    To address the redundancy and volume of raw AIS data, the observed trajectories are first transformed into a unified motion feature space. Specifically, each observed trajectory $X_i$ is shifted relative to its starting point and scaled by its overall displacement to obtain a normalized, flattened feature vector $f_i$. Using the $K$-means algorithm on these features, the trajectories are grouped into $K$ clusters, where $K = \min(K_{\text{max}}, N)$. We select a medoid-like representative trajectory for each cluster $C_k$. We compute the mean feature vector of the cluster and identify the actual trajectory whose feature is closest to this mean, i.e.,
    \begin{equation}
        i^* = \arg\min_{i \in C_k} \| f_i - \frac{1}{|C_k|} \sum_{j \in C_k} f_j \|_2.
    \end{equation}

    The representative observed and future trajectories for cluster $C_k$ are then defined as $\widetilde{X}_k = X_{i^*}$ and $\widetilde{Y}_k = Y_{i^*}$. The trajectory bank is stored as $\mathcal{B}_{\text{ais}} = \{ \widetilde{X}_k, \widetilde{Y}_k \}_{k=1}^K$.
\subsubsection{Trajectory Search and Similarity Calculation}
    For the target vessel's observed AIS trajectory $A_v$, we extract its motion feature vector $f_v$ using the same shift-and-scale normalization process. The VGTB searches $\mathcal{B}_{\text{ais}}$ to find the most similar representative trajectory using cosine similarity in the feature space, that is,
    \begin{equation}
        s_k = \frac{f_v \cdot f_k}{\|f_v\| \|f_k\| + \epsilon}, \quad k = 1, \dots, K,
    \end{equation}
    where $f_k$ is the feature vector of $\widetilde{X}_k$, $\epsilon = 10^{-8}$ ensures numerical stability, and $\cdot$ denotes the dot product. The index $k^*$ is selected as the one maximizing $s_k$.
    \begin{figure}
        \centering
        \includegraphics[width=1.00\linewidth]{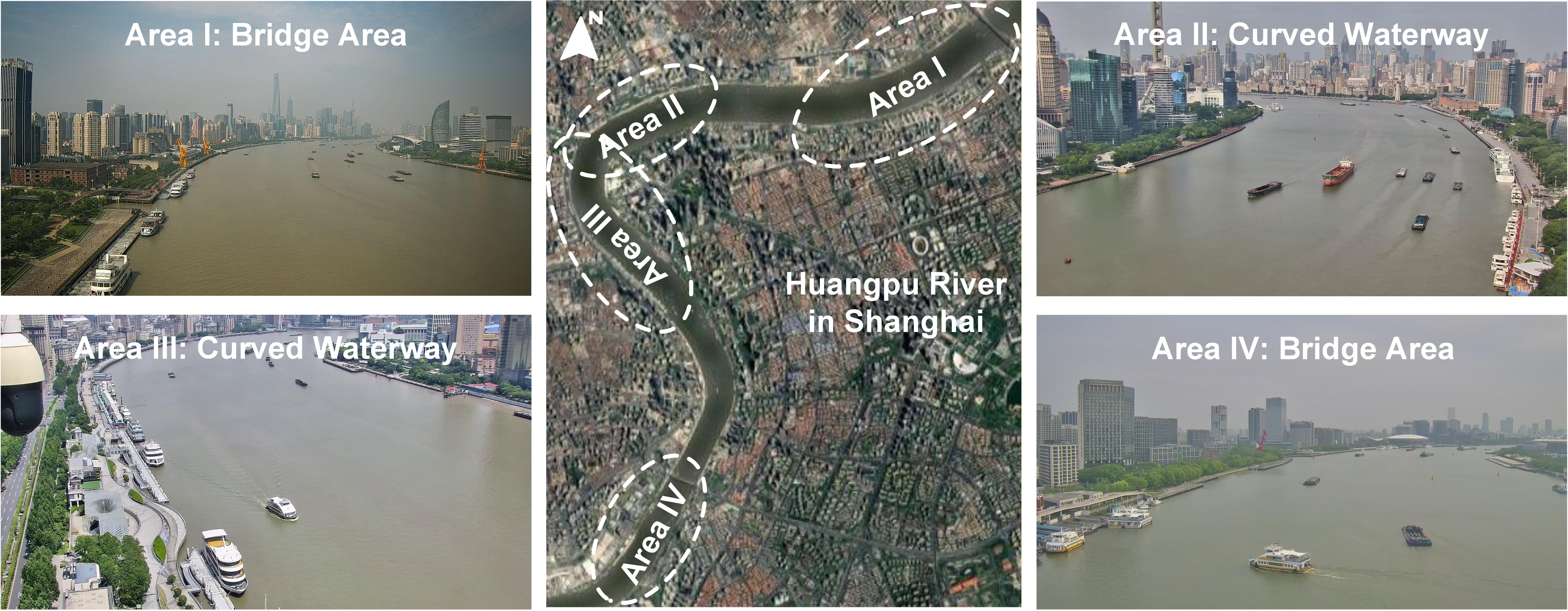}
        \caption{The proposed Maritime-MmD$^+$ focuses on four critical areas, including the bridge area and the curved waterway sections, which are characterized by their elevated safety risks and critical importance to navigational safety.}
        \label{fig:placeholder}
    \end{figure}
    \setlength{\tabcolsep}{3.00pt}
    \begin{table}[t]
        \centering
        \footnotesize
        \caption{Details of the Maritime-MmD$^+$. The "VD" represents the vessel density, with $\phi_l$, $\phi_m$, and $\phi_h$ indicating low, medium, and high densities, respectively. The “NoVA” and “NoVC” are the numbers of vessels, and the “NoPA” and “NoPC” are the numbers of vessel trajectory points in AIS and CCTV.}
        \begin{tabular}{l|cccccccc}\hline
            Clips   & Scene     & Duration & Train & VD & NoVA & NoVC & NoPA  & NoPC  \\\hline\hline
            CLIP-01 & Area I   & 80 min  & \Checkmark & ---     & 49   & 60   & 7530  & 15600  \\
            CLIP-02 & Area I   & 10 min  & \XSolidBrush  & $\phi_l$     & 15   & 21   & 1200  & 1566  \\
            CLIP-03 & Area I   & 10 min  & \XSolidBrush  & $\phi_m$     & 20   & 25   & 1447  & 1654  \\
            CLIP-04 & Area I   & 10 min  & \XSolidBrush   & $\phi_h$    & 45   & 48   & 3226  & 3429  \\\hline
            CLIP-05 & Area II   & 80 min  & \Checkmark  & ---     & 52   & 78   & 7208  & 19878  \\
            CLIP-06 & Area II   & 10 min  & \XSolidBrush  & $\phi_l$    & 16   & 29   & 1056   & 2453  \\
            CLIP-07 & Area II   & 10 min  & \XSolidBrush  & $\phi_m$     & 32   & 47   & 1847  & 3297  \\
            CLIP-08 & Area II   & 10 min  & \XSolidBrush  & $\phi_h$    & 48   & 60   & 3296  & 4463  \\\hline
            CLIP-09 & Area III   & 80 min  & \Checkmark & ---      & 44   & 55   & 3693  & 8711  \\
            CLIP-10 & Area III   & 10 min  & \XSolidBrush  & $\phi_l$    & 11   & 21   & 898   & 1644  \\
            CLIP-11 & Area III   & 10 min  & \XSolidBrush  & $\phi_m$     & 27   & 32   & 1299  & 1899  \\
            CLIP-12 & Area III   & 10 min  & \XSolidBrush  & $\phi_h$     & 38   & 44   & 2146  & 2835  \\\hline
            CLIP-13 & Area IV   & 80 min  & \Checkmark  & ---    & 136   & 169   & 17671 & 34103 \\
            CLIP-14 & Area IV   & 10 min  & \XSolidBrush & $\phi_l$      & 26  & 46  & 1376 & 3330 \\
            CLIP-15 & Area IV   & 10 min  & \XSolidBrush  & $\phi_m$     & 33   & 58   & 1772  & 4590  \\
            CLIP-16 & Area IV   & 10 min  & \XSolidBrush  & $\phi_h$    & 48   & 68   & 2941  & 5008  \\\hline
        \end{tabular}\label{table:dataset}
    \end{table}
\subsubsection{Trajectory Refinement and Fusion}
    The representative future trajectory $\widetilde{Y}_{k^*}$ is retrieved as the prior prediction $\widehat{Y}_{v,\text{prior}}^{(A)}$. Unlike directly adding an offset, the final prediction fuses this prior with a base prediction $\widehat{Y}_{v,\text{base}}^{(A)}$ generated by the main network. A learnable offset $O_v$ is computed via an MLP that takes the concatenated prior prediction and the decoded latent features as input. The refined prior is obtained as $\widehat{Y}_{v,\text{ref}}^{(A)} = \widehat{Y}_{v,\text{prior}}^{(A)} + \gamma O_v$ (where $\gamma=0.5$ is a scaling factor in our implementation). To achieve a coherent fusion, we introduce a dynamic gating network that learns an adaptive weight $\beta \in [0, 1]$ from the agent's context features. The final prediction is formulated as
    \begin{equation}
        \widehat{Y}_v^{(A)} = \beta \widehat{Y}_{v,\text{ref}}^{(A)} + (1 - \beta) \widehat{Y}_{v,\text{base}}^{(A)},
    \end{equation}
    where $\beta$ dynamically balances the contribution of the network's base prediction and the GTB-refined prior for each specific vessel.
\subsection{Loss Function}
    To ensure consistent cross-modal learning, we adopt joint supervision over AIS-based predictions $\widehat{Y}_{v}^{A}$ and CCTV-based predictions $\widehat{Y}_{v}^{C}$. Additionally, to regularize the latent distribution in our multi-modal prediction framework, we incorporate a Kullback-Leibler (KL) divergence loss. The total loss function is formulated as
    \begin{equation}
        \min_{\boldsymbol\theta}\ \mathcal{L}_{\text{total}} = \mathcal{L}_{\text{rec}} + \gamma \mathcal{L}_{\text{KL}},
    \end{equation}
    where $\mathcal{L}_{\text{rec}}$ is the reconstruction loss, $\mathcal{L}_{\text{KL}}$ is the KL divergence term, and $\gamma$ balances the reconstruction accuracy and regularization strength. We empirically set $\gamma = 0.01$.
\subsubsection{Reconstruction Loss}
    The reconstruction loss minimizes the discrepancy between the ground-truth trajectories and the best predicted mode across both modalities, i.e.,
    \begin{equation}
        \mathcal{L}_{\text{rec}} = \min_{k} \frac{1}{T_{\mathrm{fut}}} \sum_{t=1}^{T_{\mathrm{fut}}} ( \bigl\| \widehat{y}^{A}_{v,k,t} - y^{A}_{v,t} \bigr\|_2 + \bigl\| \widehat{y}^{C}_{v,k,t} - y^{C}_{v,t} \bigr\|_2 ),
    \end{equation}
   where $\widehat{y}^{A}_{v,k,t}$ and $\widehat{y}^{C}_{v,k,t}$ denote the $k$-th mode predictions at step $t$, while $y^{A}_{v,t}$ and $y^{C}_{v,t}$ are the corresponding ground truths.
    \begin{figure}
        \centering
        \includegraphics[width=1.00\linewidth]{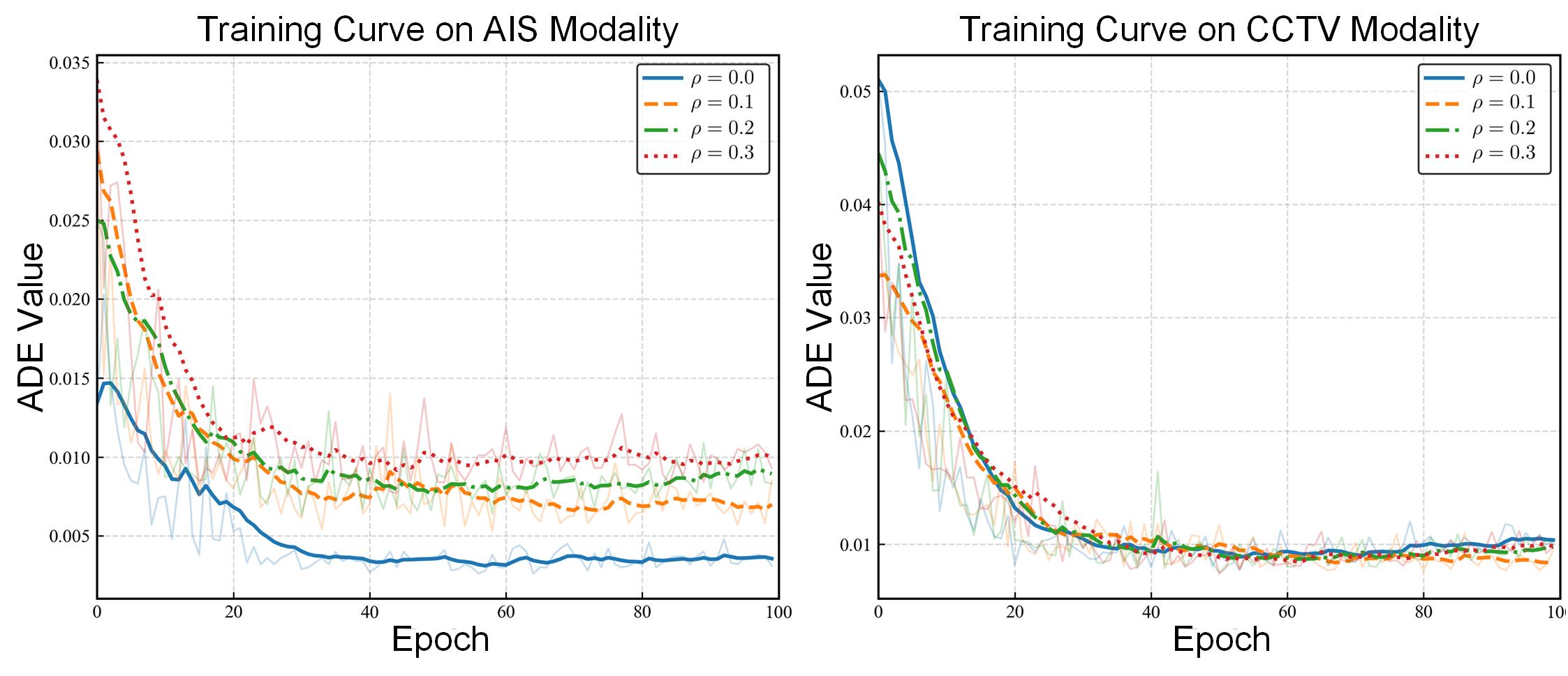}
        \caption{The AIS and CCTV convergence performance of our method under varying AIS data missing rates. Benefiting from the supplementary visual features of CCTV, our method achieves stable convergence and robust trajectory prediction despite severe AIS data loss.}
        \label{fig:convergence}
    \end{figure}
\subsubsection{KL Divergence Loss}
    To enable stochastic multi-modal predictions, the KL divergence term regularizes the learned posterior distribution by encouraging it to align with a standard normal prior. This ensures that the latent space is well-structured and supports diverse yet coherent predictions. The KL divergence loss is formulated as
    \begin{equation}
        \mathcal{L}_{\text{KL}} = -\frac{1}{2} \sum_{j=1}^{J} \left(1 + \log\sigma_j^2 - \mu_j^2 - \sigma_j^2\right),
    \end{equation}
    where $\mu_j$ and $\sigma_j$ represent the mean and standard deviation of the latent variable distribution, and $J$ is the dimensionality of the latent space. In our work, the standard deviation $\sigma_j$ is computed from the predicted log-variance $\log\sigma_j^2$ using the relationship $\sigma_j = \exp(0.5 \cdot \log\sigma_j^2)$. A latent variable $Z$ is then sampled via the reparameterization trick: $Z = \mu + \epsilon \cdot \sigma$, where $\epsilon$ is drawn from a standard normal distribution.
\section{Experiments and Discussions}\label{exper}
    This section details the experimental setup and results, including the datasets, platform, baselines, and evaluation metrics. We evaluate CmIVTP both quantitatively and qualitatively against competing methods, followed by ablation studies to validate its key architectural components. Finally, we discuss limitations and future work.
\subsection{Implementation Details}
    \setlength{\tabcolsep}{2.50pt}
    \begin{table}[t]
        \centering
        \scriptsize
        \caption{Performance of CmIVTP vs. baselines under different prediction steps. ADE and FDE ($10^{-2}$) are reported as Mean $\pm$ Std. Best in \textbf{bold}, second \underline{underlined}.}
        \begin{tabular}{l|cccccc}\hline
            & \multicolumn{2}{c}{$\Delta t = 12$}      & \multicolumn{2}{c}{$\Delta t = 24$}      & \multicolumn{2}{c}{$\Delta t = 36$}      \\
            & ADE            & FDE            & ADE            & FDE            & ADE            & FDE            \\\hline\hline
            GRU \cite{cho2014learning}              & 0.61$\spm$0.51 & 0.68$\spm$0.56 & 0.82$\spm$0.74 & 1.02$\spm$0.95 & 0.91$\spm$0.75 & \underline{1.07$\spm$0.70} \\
            BiGRU \cite{cho2014learning}            & 0.60$\spm$0.49 & 0.65$\spm$0.52 & 0.87$\spm$0.76 & 1.05$\spm$0.94 & 0.90$\spm$0.76 & 1.09$\spm$0.77 \\
            LSTM \cite{hochreiter1997long}          & 0.61$\spm$0.57 & 0.72$\spm$0.68 & 0.82$\spm$0.82 & 1.07$\spm$1.06 & 0.95$\spm$0.86 & 1.17$\spm$0.92 \\
            BiLSTM \cite{schuster1997bidirectional} & 0.61$\spm$0.53 & 0.69$\spm$0.60 & 0.86$\spm$0.84 & 1.11$\spm$1.11 & 0.96$\spm$0.84 & 1.14$\spm$0.83 \\
            TCNN \cite{bai2018empirical}            & 0.80$\spm$0.78 & 1.21$\spm$1.16 & 1.48$\spm$1.51 & 2.48$\spm$2.47 & 1.66$\spm$1.39 & 2.75$\spm$2.18 \\
            Trans. \cite{vaswani2017attention}      & \underline{0.53$\spm$0.50} & \underline{0.59$\spm$0.55} & \underline{0.71$\spm$0.77} & \underline{0.90$\spm$0.96} & \underline{0.86$\spm$0.82} & 1.09$\spm$0.87 \\
            GNN \cite{scarselli2008graph}           & 0.89$\spm$0.74 & 1.35$\spm$1.15 & 1.34$\spm$1.34 & 2.14$\spm$2.15 & 1.43$\spm$1.19 & 2.25$\spm$1.81 \\
            VAE \cite{kingma2013auto}               & 0.94$\spm$0.68 & 1.01$\spm$0.75 & 1.23$\spm$0.88 & 1.35$\spm$1.06 & 1.20$\spm$0.76 & 1.29$\spm$0.69 \\
            GAN \cite{goodfellow2020generative}     & 0.78$\spm$0.61 & 0.86$\spm$0.69 & 1.04$\spm$0.82 & 1.22$\spm$1.05 & 1.04$\spm$0.90 & 1.17$\spm$0.86 \\
            ACNet \cite{shin2024deep}               & 0.82$\spm$0.68 & 0.99$\spm$0.84 & 1.31$\spm$1.34 & 1.78$\spm$1.85 & 1.60$\spm$1.65 & 2.02$\spm$2.02 \\
            DJANet \cite{jiang2025stia}             & 0.71$\spm$0.63 & 0.84$\spm$0.75 & 0.85$\spm$0.82 & 1.05$\spm$0.99 & 0.97$\spm$0.85 & 1.21$\spm$0.91 \\
            TrAIS. \cite{nguyen2024transformer}     & 1.63$\spm$1.36 & 1.94$\spm$1.72 & 1.97$\spm$1.83 & 2.46$\spm$2.39 & 2.19$\spm$1.90 & 2.59$\spm$2.11 \\
            STGTP \cite{gong2025uncertainty}        & 0.97$\spm$0.55 & 1.04$\spm$0.56 & 1.09$\spm$0.89 & 1.31$\spm$1.17 & 1.29$\spm$1.07 & 1.47$\spm$1.18 \\\hline
            Ours                                    & \textbf{0.40$\spm$0.55} & \textbf{0.46$\spm$0.60} & \textbf{0.59$\spm$0.81} & \textbf{0.80$\spm$1.04} & \textbf{0.65$\spm$0.71} & \textbf{0.90$\spm$0.80} \\\hline
        \end{tabular}\label{table:steps}
    \end{table}
    \setlength{\tabcolsep}{2.50pt}
    \begin{table}[t]
        \centering
        \scriptsize
        \caption{Performance of CmIVTP vs. baselines under different vessel densities. ADE and FDE ($10^{-2}$) are reported as Mean $\pm$ Std. Best in \textbf{bold}, second \underline{underlined}.}
        \begin{tabular}{l|cccccc}\hline
            & \multicolumn{2}{c}{$\Phi = \phi_l$}      & \multicolumn{2}{c}{$\Phi = \phi_m$}      & \multicolumn{2}{c}{$\Phi = \phi_h$}      \\
            & ADE            & FDE            & ADE            & FDE            & ADE            & FDE            \\\hline\hline
            GRU \cite{cho2014learning}              & 3.47$\spm$5.74 & 3.49$\spm$5.14 & 0.72$\spm$0.33 & \underline{0.90$\spm$0.39} & 0.91$\spm$0.75 & \underline{1.07$\spm$0.70} \\
            BiGRU \cite{cho2014learning}            & \underline{3.10$\spm$4.82} & \underline{3.38$\spm$4.66} & 0.75$\spm$0.31 & 0.93$\spm$0.37 & 0.90$\spm$0.76 & 1.09$\spm$0.77 \\
            LSTM \cite{hochreiter1997long}          & 3.70$\spm$6.02 & 3.91$\spm$5.64 & \underline{0.70$\spm$0.32} & 1.01$\spm$0.47 & 0.95$\spm$0.86 & 1.17$\spm$0.92 \\
            BiLSTM \cite{schuster1997bidirectional} & 3.51$\spm$5.65 & 4.06$\spm$5.86 & 0.80$\spm$0.40 & 1.15$\spm$0.66 & 0.96$\spm$0.84 & 1.14$\spm$0.83 \\
            TCNN \cite{bai2018empirical}            & 3.36$\spm$4.72 & 4.12$\spm$4.92 & 1.82$\spm$1.05 & 3.30$\spm$2.48 & 1.66$\spm$1.39 & 2.75$\spm$2.18 \\
            Trans. \cite{vaswani2017attention}      & 3.28$\spm$5.55 & 3.43$\spm$5.17 & 0.70$\spm$0.39 & 1.01$\spm$0.66 & \underline{0.86$\spm$0.82} & 1.09$\spm$0.87 \\
            GNN \cite{scarselli2008graph}           & 3.34$\spm$4.76 & 4.07$\spm$4.98 & 1.76$\spm$1.37 & 2.92$\spm$2.76 & 1.43$\spm$1.19 & 2.25$\spm$1.81 \\
            VAE \cite{kingma2013auto}               & 4.78$\spm$7.68 & 4.57$\spm$6.67 & 0.98$\spm$0.37 & 1.13$\spm$0.38 & 1.20$\spm$0.76 & 1.29$\spm$0.69 \\
            GAN \cite{goodfellow2020generative}     & 4.35$\spm$7.18 & 4.30$\spm$6.50 & 0.87$\spm$0.39 & 1.05$\spm$0.55 & 1.04$\spm$0.90 & 1.17$\spm$0.86 \\
            ACNet \cite{shin2024deep}               & 5.29$\spm$8.46 & 5.25$\spm$7.63 & 1.27$\spm$0.68 & 1.68$\spm$1.09 & 1.60$\spm$1.65 & 2.02$\spm$2.02 \\
            DJANet \cite{jiang2025stia}             & 4.01$\spm$6.75 & 3.52$\spm$5.00 & 0.72$\spm$0.36 & 0.90$\spm$0.51 & 0.97$\spm$0.85 & 1.21$\spm$0.91 \\
            TrAIS. \cite{nguyen2024transformer}     & 6.09$\spm$9.22 & 5.60$\spm$7.55 & 1.70$\spm$1.48 & 2.06$\spm$1.90 & 2.19$\spm$1.90 & 2.59$\spm$2.11 \\
            STGTP \cite{gong2025uncertainty}        & 4.25$\spm$6.57 & 4.16$\spm$5.79 & 1.55$\spm$1.13 & 1.87$\spm$1.51 & 1.29$\spm$1.07 & 1.47$\spm$1.18 \\\hline
            Ours                                    & \textbf{2.24$\spm$3.40} & \textbf{3.22$\spm$4.50} & \textbf{0.68$\spm$0.54} & \textbf{0.81$\spm$0.59} & \textbf{0.65$\spm$0.71} & \textbf{0.90$\spm$0.80} \\\hline
            \end{tabular}\label{table:density}
    \end{table}
    \setlength{\tabcolsep}{2.50pt}
    \begin{table}[t]
        \centering
        \scriptsize
        \caption{Performance of CmIVTP vs. baselines under different AIS missing rates. ADE and FDE ($10^{-2}$) are reported as Mean $\pm$ Std. Best in \textbf{bold}, second \underline{underlined}.}
        \begin{tabular}{l|cccccc}\hline
            & \multicolumn{2}{c}{$\rho = 0.1$}       & \multicolumn{2}{c}{$\rho = 0.2$}       & \multicolumn{2}{c}{$\rho = 0.3$}       \\
            & ADE            & FDE            & ADE            & FDE            & ADE            & FDE            \\\hline\hline
            GRU \cite{cho2014learning}              & 1.59$\spm$1.32 & 1.61$\spm$1.27 & 1.81$\spm$1.59 & 1.85$\spm$1.54 & 1.92$\spm$1.71 & 1.94$\spm$1.70 \\
            BiGRU \cite{cho2014learning}            & 1.57$\spm$1.27 & 1.65$\spm$1.32 & 1.73$\spm$1.51 & 1.82$\spm$1.54 & 1.83$\spm$1.53 & 1.87$\spm$1.52 \\
            LSTM \cite{hochreiter1997long}          & 1.34$\spm$1.01 & 1.48$\spm$1.02 & \underline{1.53$\spm$1.20} & 1.72$\spm$1.27 & 1.88$\spm$1.67 & 1.91$\spm$1.54 \\
            BiLSTM \cite{schuster1997bidirectional} & 1.33$\spm$0.98 & 1.45$\spm$0.99 & \underline{1.53$\spm$1.20} & \underline{1.59$\spm$1.17} & 1.80$\spm$1.54 & 1.84$\spm$1.45 \\
            TCNN \cite{bai2018empirical}            & 1.74$\spm$1.40 & 2.98$\spm$2.20 & 2.00$\spm$1.52 & 3.20$\spm$2.27 & 2.29$\spm$1.71 & 3.56$\spm$2.51 \\
            Trans. \cite{vaswani2017attention}      & 1.45$\spm$1.15 & 1.48$\spm$1.09 & 1.78$\spm$1.62 & 1.83$\spm$1.64 & 1.99$\spm$1.92 & 1.92$\spm$1.70 \\
            GNN \cite{scarselli2008graph}           & 1.56$\spm$1.23 & 2.21$\spm$1.89 & 2.00$\spm$1.46 & 2.29$\spm$1.64 & 2.20$\spm$1.67 & 2.60$\spm$1.96 \\
            VAE \cite{kingma2013auto}               & 1.78$\spm$1.14 & 1.84$\spm$1.10 & 2.01$\spm$1.43 & 2.10$\spm$1.42 & 1.88$\spm$1.39 & 1.90$\spm$1.29 \\
            GAN \cite{goodfellow2020generative}     & 1.59$\spm$1.15 & 1.63$\spm$1.09 & 1.83$\spm$1.40 & 1.87$\spm$1.36 & 1.81$\spm$1.40 & 1.86$\spm$1.38 \\
            ACNet \cite{shin2024deep}               & 1.99$\spm$1.57 & 2.37$\spm$1.89 & 2.17$\spm$1.60 & 2.47$\spm$1.84 & 2.19$\spm$1.62 & 2.52$\spm$1.95 \\
            DJANet \cite{jiang2025stia}             & \underline{1.22$\spm$0.88} & \underline{1.28$\spm$0.86} & 1.95$\spm$1.78 & 2.05$\spm$1.80 & 2.08$\spm$2.00 & 2.19$\spm$2.09 \\
            TrAIS. \cite{nguyen2024transformer}     & 2.40$\spm$2.02 & 2.75$\spm$2.30 & 2.52$\spm$2.13 & 2.86$\spm$2.41 & 2.58$\spm$2.16 & 2.91$\spm$2.40 \\
            STGTP \cite{gong2025uncertainty}        & 1.45$\spm$1.08 & 1.56$\spm$1.08 & 1.63$\spm$1.27 & 1.72$\spm$1.26 & \underline{1.58$\spm$1.18} & \underline{1.63$\spm$1.11} \\\hline
            Ours                                    & \textbf{0.80$\spm$0.73} & \textbf{0.97$\spm$0.79} & \textbf{0.90$\spm$0.73} & \textbf{1.03$\spm$0.76} & \textbf{1.08$\spm$0.88} & \textbf{1.22$\spm$0.93} \\\hline
        \end{tabular}\label{table:miss}
    \end{table}
    \setlength{\tabcolsep}{4.75pt}
    \begin{table*}[t]
        \centering
        \scriptsize
        \caption{Performance of CmIVTP vs. baselines for AIS and CCTV trajectory prediction using no-missing and missing AIS data. ADE and FDE ($10^{-2}$) are reported as Mean $\pm$ Std. Best in \textbf{bold}, second \underline{underlined}.}
        \begin{tabular}{l|cccc|cccc|cccc}\hline
            & \multicolumn{4}{c|}{Prediction Step}                                          & \multicolumn{4}{c|}{Vessel Density}                                       & \multicolumn{4}{c}{AIS Missing}                                       \\
            & \multicolumn{2}{c}{AIS}         & \multicolumn{2}{c|}{CCTV}        & \multicolumn{2}{c}{AIS}         & \multicolumn{2}{c|}{CCTV}        & \multicolumn{2}{c}{AIS}         & \multicolumn{2}{c}{CCTV}        \\
            & ADE            & FDE            & ADE            & FDE            & ADE            & FDE            & ADE            & FDE            & ADE            & FDE            & ADE            & FDE            \\\hline\hline
            GRU \cite{cho2014learning}              & 0.75$\spm$0.55 & 0.92$\spm$0.67 & 2.24$\spm$1.57 & 3.32$\spm$2.26 & 1.41$\spm$2.80 & 1.61$\spm$2.54 & 4.34$\spm$5.35 & 6.63$\spm$6.29 & 1.83$\spm$1.39 & 1.85$\spm$1.37 & 2.64$\spm$1.73 & 4.64$\spm$3.10 \\
            BiGRU \cite{cho2014learning}            & 0.74$\spm$0.63 & 0.86$\spm$0.73 & \underline{1.85$\spm$1.29} & \underline{2.79$\spm$1.91} & 1.39$\spm$2.49 & 1.59$\spm$2.46 & 3.84$\spm$5.35 & 5.86$\spm$6.08 & 1.71$\spm$1.29 & 1.78$\spm$1.30 & 2.50$\spm$1.61 & 4.42$\spm$2.88 \\
            LSTM \cite{hochreiter1997long}          & 0.74$\spm$0.65 & 0.92$\spm$0.76 & 2.28$\spm$1.69 & 3.37$\spm$2.36 & 1.78$\spm$3.36 & 2.03$\spm$3.20 & 4.50$\spm$5.56 & 6.70$\spm$6.30 & 1.58$\spm$1.19 & 1.70$\spm$1.17 & 2.66$\spm$1.87 & 4.67$\spm$3.02 \\
            BiLSTM \cite{schuster1997bidirectional} & 0.75$\spm$0.61 & 0.95$\spm$0.78 & 2.42$\spm$1.92 & 3.67$\spm$2.90 & 1.73$\spm$3.25 & 2.12$\spm$3.42 & 4.98$\spm$6.67 & 7.70$\spm$8.18 & 1.56$\spm$1.14 & \underline{1.63$\spm$1.10} & 2.75$\spm$2.03 & 4.60$\spm$3.19 \\
            TCNN \cite{bai2018empirical}            & 1.28$\spm$1.11 & 2.09$\spm$1.78 & 3.54$\spm$1.70 & 7.17$\spm$3.58 & 1.70$\spm$1.76 & 2.78$\spm$2.36 & 4.36$\spm$2.41 & 8.77$\spm$3.24 & 2.01$\spm$1.38 & 3.24$\spm$2.06 & 4.53$\spm$1.58 & 10.53$\spm$3.63 \\
            Trans. \cite{vaswani2017attention}      & \underline{0.69$\spm$0.61} & \underline{0.85$\spm$0.71} & 1.96$\spm$1.46 & 3.06$\spm$2.37 & 1.61$\spm$3.06 & 1.84$\spm$2.89 & 4.04$\spm$4.96 & 6.56$\spm$6.26 & 1.74$\spm$1.43 & 1.75$\spm$1.35 & \underline{2.29$\spm$1.53} & \underline{4.07$\spm$2.78} \\
            GNN \cite{scarselli2008graph}           & 1.12$\spm$0.88 & 1.80$\spm$1.40 & 3.19$\spm$1.69 & 5.76$\spm$3.16 & 2.30$\spm$2.80 & 3.25$\spm$3.18 & 4.56$\spm$3.08 & 8.02$\spm$3.94 & 1.69$\spm$1.31 & 2.03$\spm$1.50 & 3.25$\spm$1.62 & 6.54$\spm$2.99 \\
            VAE \cite{kingma2013auto}               & 1.07$\spm$0.70 & 1.18$\spm$0.81 & 3.21$\spm$1.30 & 4.27$\spm$1.73 & 2.73$\spm$4.90 & 2.71$\spm$4.29 & 5.15$\spm$3.95 & 7.30$\spm$4.49 & 1.88$\spm$1.23 & 1.94$\spm$1.20 & 3.78$\spm$1.75 & 6.01$\spm$2.81 \\
            GAN \cite{goodfellow2020generative}     & 0.96$\spm$0.71 & 1.08$\spm$0.80 & 2.72$\spm$1.70 & 3.74$\spm$2.35 & 2.16$\spm$4.11 & 2.25$\spm$3.76 & 4.49$\spm$4.27 & 6.55$\spm$5.10 & 1.74$\spm$1.18 & 1.79$\spm$1.16 & 3.13$\spm$2.14 & 5.10$\spm$3.51 \\
            ACNet \cite{shin2024deep}               & 1.25$\spm$1.18 & 1.60$\spm$1.53 & 3.52$\spm$2.54 & 5.07$\spm$3.80 & 2.85$\spm$4.89 & 3.10$\spm$4.49 & 4.91$\spm$4.54 & 7.11$\spm$4.84 & 2.13$\spm$1.50 & 2.47$\spm$1.78 & 4.35$\spm$3.03 & 7.10$\spm$4.73 \\
            DJANet \cite{jiang2025stia}             & 0.95$\spm$0.72 & 1.17$\spm$0.83 & 2.59$\spm$1.69 & 4.10$\spm$2.67 & \underline{1.36$\spm$2.72} & \underline{1.47$\spm$2.08} & \underline{3.35$\spm$3.20} & \underline{5.64$\spm$4.30} & 1.89$\spm$1.47 & 1.98$\spm$1.52 & 2.75$\spm$1.66 & 4.79$\spm$2.82 \\
            TrAIS. \cite{nguyen2024transformer}     & 2.12$\spm$1.48 & 2.56$\spm$1.81 & 5.89$\spm$3.73 & 7.62$\spm$4.72 & 3.85$\spm$5.49 & 3.93$\spm$4.56 & 7.79$\spm$5.57 & 10.25$\spm$6.71 & 2.58$\spm$1.83 & 2.92$\spm$2.07 & 6.69$\spm$4.39 & 8.90$\spm$5.48 \\
            STGTP \cite{gong2025uncertainty}        & 1.06$\spm$0.78 & 1.19$\spm$0.89 & 2.78$\spm$1.93 & 3.85$\spm$2.84 & 2.36$\spm$3.66 & 2.50$\spm$3.30 & 5.68$\spm$5.95 & 7.94$\spm$6.80 & \underline{1.55$\spm$1.08} & 1.64$\spm$1.07 & 3.04$\spm$1.91 & 5.03$\spm$3.40 \\\hline
            Ours                                    & \textbf{0.55$\spm$0.62} & \textbf{0.72$\spm$0.76} & \textbf{0.61$\spm$0.46} & \textbf{1.20$\spm$1.11} & \textbf{0.74$\spm$1.05} & \textbf{1.21$\spm$1.99} & \textbf{1.41$\spm$1.20} & \textbf{3.31$\spm$3.03} & \textbf{0.97$\spm$0.69} & \textbf{1.13$\spm$0.73} & \textbf{0.86$\spm$0.58} & \textbf{1.75$\spm$1.26} \\\hline
        \end{tabular}\label{table:aisvideo}
    \end{table*}
\subsubsection{Optimizer and Experimental Setup}
    The Adam optimizer is employed with a learning rate of 0.0001. The batch size is dynamically defined by the number of trajectories per batch, and the model is trained for 100 epochs. The implementation is developed using the PyTorch framework. The learning rate is adjusted during training using the \texttt{ReduceLROnPlateau} scheduler, which reduces the learning rate by a factor of 0.5 when the training loss plateaus for 10 consecutive epochs. All experiments are executed on an Ubuntu 18.04 operating system equipped with an NVIDIA RTX 4090 GPU, utilizing CUDA for accelerated computation.
\subsubsection{Datasets}
    In this work, we present an enhanced version of the Maritime-MmD dataset \cite{lu2026graph}, now named Maritime-MmD$^+$. It expands and standardizes its applicability for maritime-specific tasks such as vessel detection, trajectory tracking and prediction, data fusion, and scene understanding. As shown in Table \ref{table:dataset} and Fig. \ref{fig:placeholder}, the dataset includes the original $16$ AIS data and CCTV videos collected along the Huangpu River in Shanghai, China, capturing a wide range of scenarios such as bridge regions and curved waterways. To comprehensively evaluate the model's robustness, generalization, and resilience to data incompleteness within these scenarios, our experiments explicitly analyze performance across varying time steps ($\Delta t = 12, 24, 36$), diverse vessel mixes represented by different density levels ($\Phi = \phi_l, \phi_m, \phi_h$), and simulated AIS missing rates ($\rho = 0.1, 0.2, 0.3$). Notably, $\rho$ represents the proportion of 'dark vessels' completely lacking AIS data, rather than intermittent dropouts.
\subsubsection{Data preprocessing and Alignment}
    To ensure meaningful cross-modal interaction, the heterogeneous AIS and processed CCTV video streams \cite{guo2023asynchronous,liu2024real} are first aligned in both temporal and spatial dimensions. Given the discrepancies between the two modalities in sampling frequency, coordinate systems, and observation noise, we synchronize AIS trajectories with CCTV frames through temporal resampling and estimation, while handling spatial inconsistency using the multi-vessel association framework established in our prior work \cite{lu2026uncertainty,lu2026graph}. By leveraging graph learning, this framework robustly associates AIS kinematic sequences with their corresponding CCTV visual targets under heterogeneous observations, thereby alleviating the effects of coordinate discrepancies and noise before multimodal fusion.
\subsubsection{Competitive Methods}
    To evaluate the performance of our proposed method, we conducted comparisons against a range of baseline predictive models and recently proposed task-specific methods, covering both deterministic and uncertainty-aware prediction. The baseline models included gated recurrent units (GRU) \cite{cho2014learning}, bidirectional GRU (Bi-GRU) \cite{cho2014learning}, long short-term memory (LSTM) networks \cite{hochreiter1997long}, bidirectional LSTM (Bi-LSTM) \cite{schuster1997bidirectional}, temporal convolutional neural networks (TCNN) \cite{bai2018empirical}, Transformer (Trans.) \cite{vaswani2017attention}, group neural network (GNN) \cite{scarselli2008graph}, variational autoencoder (VAE) \cite{kingma2013auto}, and generative adversarial network (GAN) \cite{goodfellow2020generative}. The recently proposed methods included ACNet \cite{shin2024deep}, DJANet \cite{jiang2025stia}, TrAIS \cite{nguyen2024transformer}, and STGTP \cite{gong2025uncertainty}.
    
    All compared methods were trained and evaluated on the same dataset using identical train/validation/test splits, evaluation metrics, and training protocols. For the main comparison, all baselines used the same multimodal inputs, i.e., AIS trajectories and CCTV trajectories. Our model further incorporates scene representations as an additional visual semantic cue. Moreover, each baseline was implemented and tuned following its original design. To ensure the statistical significance and stability of our evaluation, all quantitative experiments are independently repeated 10 times using different random seeds. The final results are reported as the mean and standard deviation across these 10 runs.
\subsubsection{Performance Evaluation}
    The performance of trajectory prediction is evaluated using two widely adopted metrics, that is, average displacement error (ADE) and final displacement error (FDE). ADE measures the average Euclidean distance between predicted trajectories and ground truth trajectories across all predicted time steps, while FDE evaluates the Euclidean distance between the predicted final positions and the ground truth final destinations. For both metrics, lower values indicate better prediction performance.
    \begin{figure*}[ht]
        \centering
        \setlength{\abovecaptionskip}{-0.25cm}
        \includegraphics[width=1.00\linewidth]{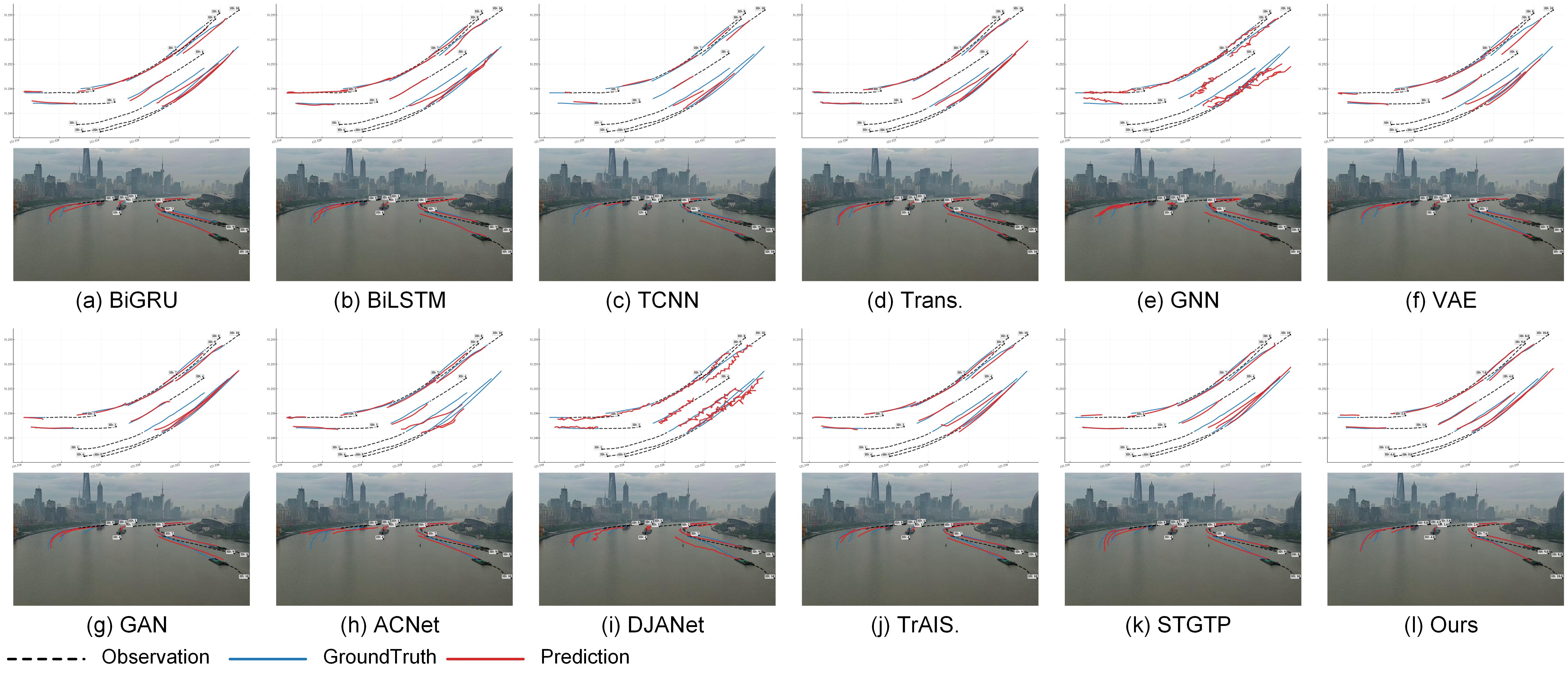}
        \caption{Qualitative comparison of trajectory prediction methods on the 36-step task using no-missing AIS data. From left to right, the images include (a) Bi-GRU \cite{cho2014learning}, (b) Bi-LSTM \cite{schuster1997bidirectional}, (c) TCNN \cite{bai2018empirical}, (d) Transformer \cite{vaswani2017attention}, (e) GNN \cite{gan2014goodfellow}, (f) VAE \cite{kingma2013auto}, (g) GAN \cite{goodfellow2020generative}, (h) ACNet \cite{shin2024deep}, (i) DJANet \cite{jiang2025stia}, (j) TrAIS. \cite{nguyen2024transformer}, (k) STGTP \cite{gong2025uncertainty}, and (l) our method, respectively.}
        \label{figure:Figure_tp21}
    \end{figure*}
    \begin{figure*}[ht]
        \centering
        \setlength{\abovecaptionskip}{-0.25cm}
        \includegraphics[width=1.00\linewidth]{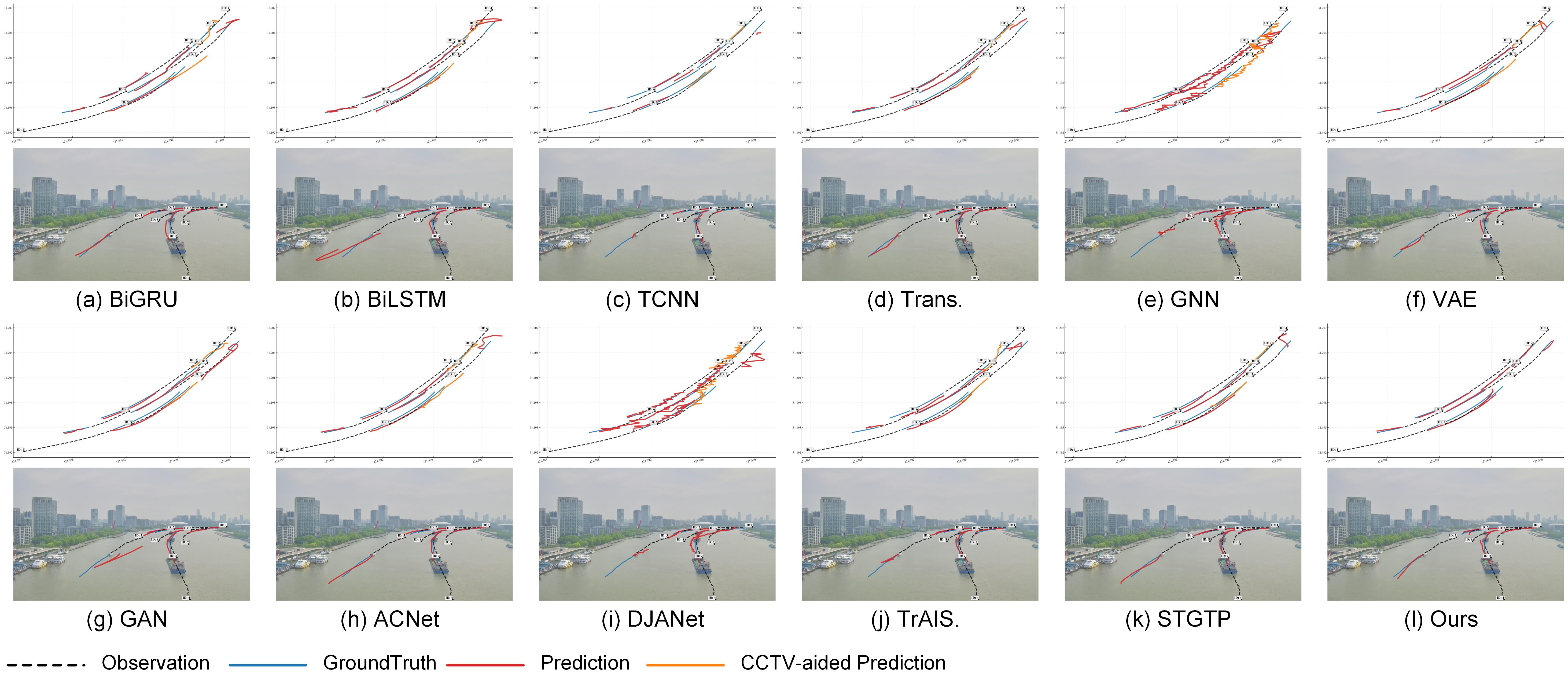}
        \caption{Qualitative comparison of trajectory prediction methods on the 36-step task using missing AIS data. We manually remove AIS trajectories and use information provided by CCTV to assist in predicting the missing AIS trajectories, which are represented with orange lines. From left to right, the images include (a) Bi-GRU \cite{cho2014learning}, (b) Bi-LSTM \cite{schuster1997bidirectional}, (c) TCNN \cite{bai2018empirical}, (d) Transformer \cite{vaswani2017attention}, (e) GNN \cite{gan2014goodfellow}, (f) VAE \cite{kingma2013auto}, (g) GAN \cite{goodfellow2020generative}, (h) ACNet \cite{shin2024deep}, (i) DJANet \cite{jiang2025stia}, (j) TrAIS. \cite{nguyen2024transformer}, (k) STGTP \cite{gong2025uncertainty}, and (l) our method, respectively.}
        \label{figure:Figure_tp22}
    \end{figure*}
    \begin{figure}[t]
        \centering
        \setlength{\abovecaptionskip}{-0.25cm}
        \includegraphics[width=1.00\linewidth]{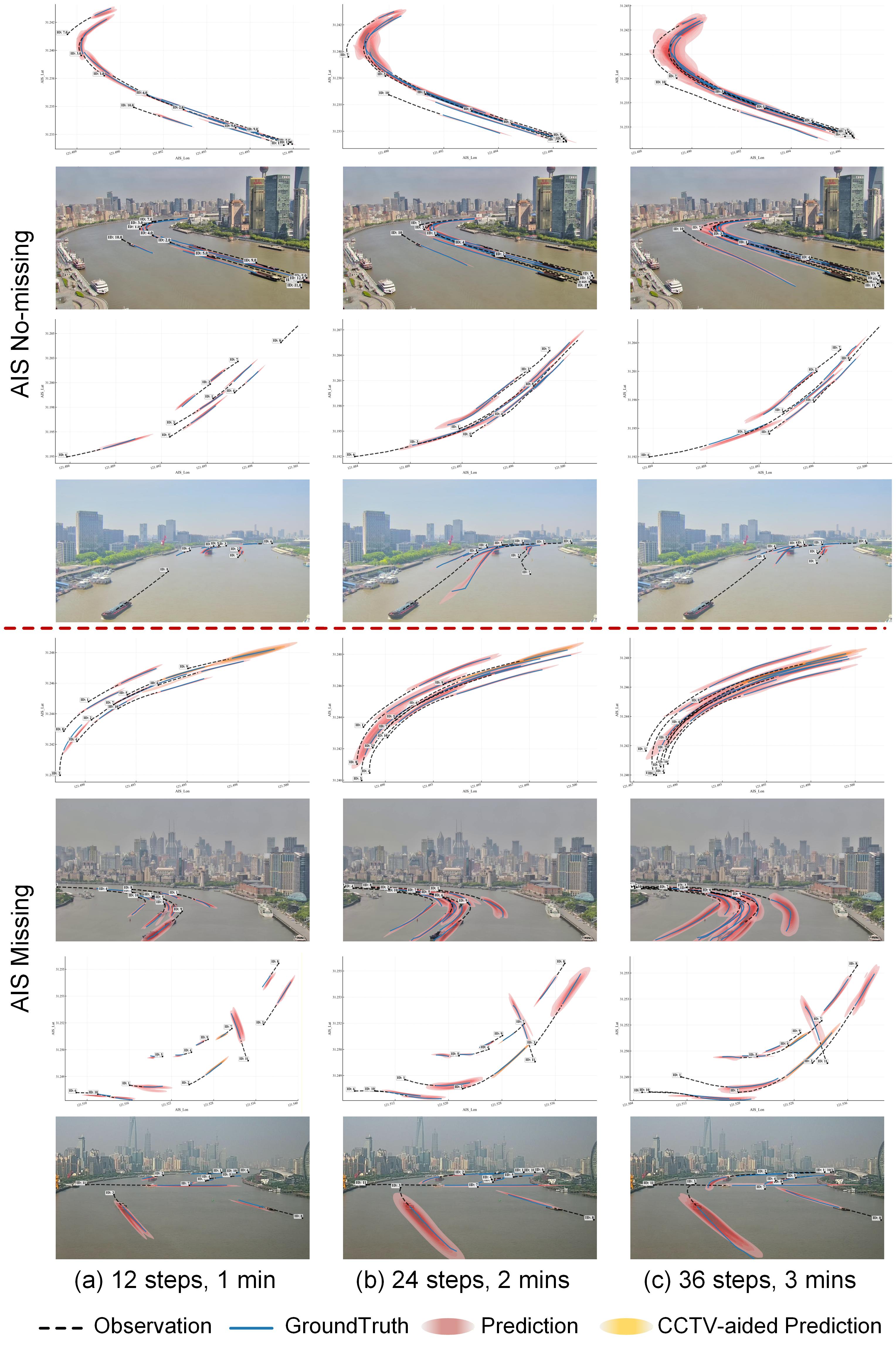}
        \caption{Qualitative comparison of CmIVTP at 12, 24, and 36 prediction steps (1, 2, and 3 minutes) under complete and missing AIS conditions. With complete AIS data, the predicted distributions are compact and closely aligned with the ground truth. In missing AIS scenarios, the model produces more diverse predictions with increased variance to capture uncertainty, while the ground truth remains well covered, demonstrating adaptive risk-aware prediction and robustness.}
        \label{figure:Figure_un}
    \end{figure}
\subsection{Quantitative Performance Analysis}
    This subsection analyzes the quantitative performance of trajectory prediction experiments, covering convergence, time step variations, density levels, and missing rates.
\subsubsection{Convergence Analysis}
    The convergence results, obtained under fixed conditions of $\Delta t = 36$ and $\Phi = \phi_h$, demonstrate that the proposed model effectively handles the negative impact of AIS data missing rates on trajectory prediction. As shown in Fig. \ref{fig:convergence}, the model achieves consistent and stable convergence across varying missing rates ($\rho = 0.0$, $\rho = 0.1$, $\rho = 0.2$, $\rho = 0.3$). While higher missing rates inevitably lead to a minor increase in the ADE on the AIS modality, the model still achieves consistent convergence. Notably, the convergence curves on the CCTV modality remain highly stable and tightly clustered regardless of the AIS missing rate. This clearly indicates that by benefiting from the supplementary visual features provided by CCTV, the model maintains robust overall performance, ensuring reliable trajectory predictions even with significant AIS data loss.
\subsubsection{Effect of Time Step Variation}
    Table \ref{table:steps} presents the comparative results of trajectory prediction across varying prediction horizons (i.e., $\Delta t = 12, 24, 36$). It is observed that predictive performance generally declines for all models as the time steps increase, primarily due to the accumulation of errors over longer sequences. However, CmIVTP consistently exhibits the lowest ADE and FDE with relatively low prediction variance across all four areas, demonstrating superior stability. For instance, traditional sequential models like TCNN and RNN experience a sharp increase in error at $\Delta t = 36$. Similarly, generative models like VAE and GAN, and recent advanced methods like ACNet, DJANet, STGTP, and TrAIS still suffer significant degradation over extended horizons despite incorporating complex spatial-temporal or probabilistic mechanisms. In contrast, CmIVTP maintains a substantially lower error margin, effectively mitigating the error propagation typical of recursive prediction.
\subsubsection{Effect of Density Variation}
    Table \ref{table:density} illustrates the impact of different vessel traffic density ($\phi_l$, $\phi_m$, and $\phi_h$) on prediction accuracy. High-density scenarios introduce complex vessel-to-vessel interactions, which typically challenge conventional models that treat trajectories in isolation. As shown in the results, both traditional sequential models and recent advanced learning-based methods exhibit pronounced performance degradation under high-density conditions, despite the latter's integration of complex spatial-temporal or probabilistic mechanisms. Conversely, CmIVTP consistently yields the lowest ADE and FDE across all density levels. By explicitly modeling inter-vessel dependencies, our method maintains robust predictive accuracy in congested waters, effectively capturing non-linear trajectory deviations induced by collision avoidance maneuvers.
\subsubsection{Effect of Missing Rate Variation} 
    Table \ref{table:miss} demonstrates the model's resilience to data incompleteness by simulating varying AIS data missing rates ($\rho = 0.1, 0.2, 0.3$). As the missing rate increases, both traditional sequential models and recent advanced methods exhibit substantial performance degradation, indicating a heavy reliance on continuous, high-quality input streams. Conversely, CmIVTP demonstrates exceptional robustness, maintaining significantly lower ADE and FDE with minimal variance even when 30\% of the data is missing ($\rho = 0.3$). Notably, the performance gap between CmIVTP and the best-performing baselines widens considerably as data sparsity increases. This superior resilience is primarily attributed to the effective utilization of multimodal features (e.g., integrating CCTV data), which successfully compensates for the incomplete AIS signals. Consequently, CmIVTP proves to be a more practical solution for real-world maritime environments where signal loss and data interruptions are frequent occurrences.
\subsubsection{Analysis of Multimodal Prediction} 
    Table \ref{table:aisvideo} presents an evaluation of CmIVTP against baselines across varying prediction steps, traffic densities, and data missingness scenarios for both AIS and CCTV modalities. Notably, while recent advanced methods maintain reasonable accuracy on standard AIS records, they suffer from severe performance degradation when predicting CCTV trajectories or handling missing data. In contrast, CmIVTP consistently achieves the lowest ADE and FDE across all configurations. The performance margin is particularly pronounced in the CCTV and missing data scenarios, where our method reduces errors substantially compared to the best-performing baselines. This exceptional robustness is driven by the CmIE module, which integrates heterogeneous AIS and CCTV features to capture complex vessel-environment dynamics and compensate for incomplete single-source inputs. Furthermore, the UaVD module explicitly models the stochastic nature of vessel movements, further enhancing predictive stability. Collectively, these results validate that CmIVTP's multimodal fusion framework successfully overcomes the inherent limitations of isolated data sources, ensuring reliable trajectory predictions even in complex, data-sparse maritime environments.
\subsection{Qualitative Performance Analysis}
    To assess the model's robustness and generalization within complex channel constraints, we visually compare our proposed method against the competitive baselines from the quantitative experiments. The evaluation focuses on two representative scenarios: AIS data complete (Fig. \ref{figure:Figure_tp21}) and AIS data missing (Fig. \ref{figure:Figure_tp22}). In the data-complete scenario, while benchmark models capture fundamental motion patterns, several baselines exhibit noticeable noise and fluctuations. Our method, however, attains superior trajectory fitting precision and smoothness through the deep fusion of multi-source heterogeneous information. Conversely, in the extreme case of AIS signal loss, unimodal baselines exhibit significant trajectory divergence, erratic fluctuations, or latency due to the interruption of critical data. By contrast, our method maintains highly stable prediction performance across both scenarios by leveraging the synergy between visual and positional data. Specifically, CCTV-aided visual features provide robust spatiotemporal compensation for AIS signal gaps, ensuring our predictions closely align with the ground truth.
\subsection{Uncertainty Prediction Analysis}
    To better evaluate CmIVTP's capability in modeling stochasticity, we compared the predicted distributions under scenarios with complete versus missing AIS data. As shown in Fig. \ref{figure:Figure_un}, with complete data, the generated trajectory distribution is compact and tightly concentrated around the ground truth, reflecting the model's high confidence. Conversely, in missing data scenarios, the predicted distribution exhibits reasonable diversity and expanded variance to capture the uncertainty arising from information loss. Despite this increased variance, the ground truth remains robustly within the prediction range. It indicates that CmIVTP can adaptively quantify prediction risk based on data quality, demonstrating strong robustness in complex waterway environments.
\subsection{Ablation Analysis}
    \setlength{\tabcolsep}{4.85pt}
    \begin{table}[t]
        \centering
        \footnotesize
        \caption{Ablation study on multimodal features.}
        \begin{tabular}{ccc|cccc}
        \hline
            \multirow{2}{*}{\makecell{AIS \\ Traj.}}  & \multirow{2}{*}{\makecell{CCTV \\ Traj.}} & \multirow{2}{*}{\makecell{Scene \\ Rep.}}  & \multicolumn{2}{c}{No Missing} & \multicolumn{2}{c}{Missing}\\
            &  &  &  ADE  & FDE & ADE  & FDE \\\hline\hline
            \Checkmark              &                &                    & 0.77$\spm$0.81 & 0.87$\spm$0.71   & --- & ---       \\
            \Checkmark              & \Checkmark               &                    & 0.64$\spm$0.73 & 0.79$\spm$0.67  & 1.27$\spm$1.05 & 1.56$\spm$1.11\\
            \Checkmark              & \Checkmark               & \Checkmark                   & 0.61$\spm$0.75 & 0.74$\spm$0.66  & 0.94$\spm$0.82 & 1.05$\spm$0.89\\\hline
        \end{tabular}\label{table:mf}
    \end{table}
    \setlength{\tabcolsep}{2.35pt}
    \begin{table}[t]
        \centering
        \footnotesize
        \caption{Ablation study on network modules.}
        \begin{tabular}{cccc|cccc}
        \hline  
        \multirow{2}{*}{VSTaE}   & \multirow{2}{*}{CmIE}   & \multirow{2}{*}{UaVD}   & \multirow{2}{*}{VGTB}   & \multicolumn{2}{c}{No Missing} & \multicolumn{2}{c}{Missing}                         \\
        &  &  &  & ADE  & FDE & ADE  & FDE \\\hline\hline
         &  \Checkmark    &      &  &   0.80$\spm$0.91     & 0.89$\spm$0.87 & 1.45$\spm$1.51   & 1.95$\spm$1.77     \\
        \Checkmark          & \Checkmark     &      &         & 0.72$\spm$0.88 & 0.86$\spm$0.92   & 1.17$\spm$1.05  & 1.33$\spm$1.21  \\
        \Checkmark          &  \Checkmark    &  \Checkmark    &  &        0.65$\spm$0.84 & 0.79$\spm$0.89   & 1.12$\spm$0.95  &  1.35$\spm$1.09 \\
        \Checkmark          &  \Checkmark    &  \Checkmark    &  \Checkmark       & 0.61$\spm$0.75 & 0.74$\spm$0.66  & 0.94$\spm$0.82 & 1.05$\spm$0.89     \\\hline
        \end{tabular}\label{as:nm}
    \end{table}
\subsubsection{Multimodal Features}
    Based on Table \ref{table:mf}, we quantitatively evaluate the contribution of different modal features. The baseline relying solely on AIS data demonstrates limited accuracy and performs poorly in data-missing scenarios. Incorporating CCTV trajectories significantly reduces errors and mitigates AIS signal loss, verifying the critical complementarity of visual features. Furthermore, integrating Scene Representations yields the best performance across all scenarios. These findings indicate that the environmental semantics extracted by VSTaE effectively capture vessel-environment constraints, demonstrating that multimodal fusion is essential for robust and accurate trajectory prediction.
\subsubsection{Network Modules}
    To verify the effectiveness of the CmIVTP architecture, we conducted ablation studies as shown in Table \ref{as:nm}. Specifically, the baseline CmIE aligns kinematic data with visual imagery via attention to handle heterogeneous fusion and capture interaction intentions. Adding VSTaE extracts global scene semantics to ensure navigation compliance and mitigate environmentally infeasible predictions. Furthermore, UaVD employs a probabilistic generation paradigm to model motion stochasticity, enhancing robustness under uncertainty. Finally, VGTB utilizes non-parametric retrieval to introduce historical priors, compensating for data sparsity. The progressive integration of these modules ensures high prediction accuracy while adapting to complex physical environments and regulations.
\subsubsection{Analysis of the UaVD}
    \begin{figure}[t]
        \centering
        \setlength{\abovecaptionskip}{-0.25cm}
        \includegraphics[width=1.00\linewidth]{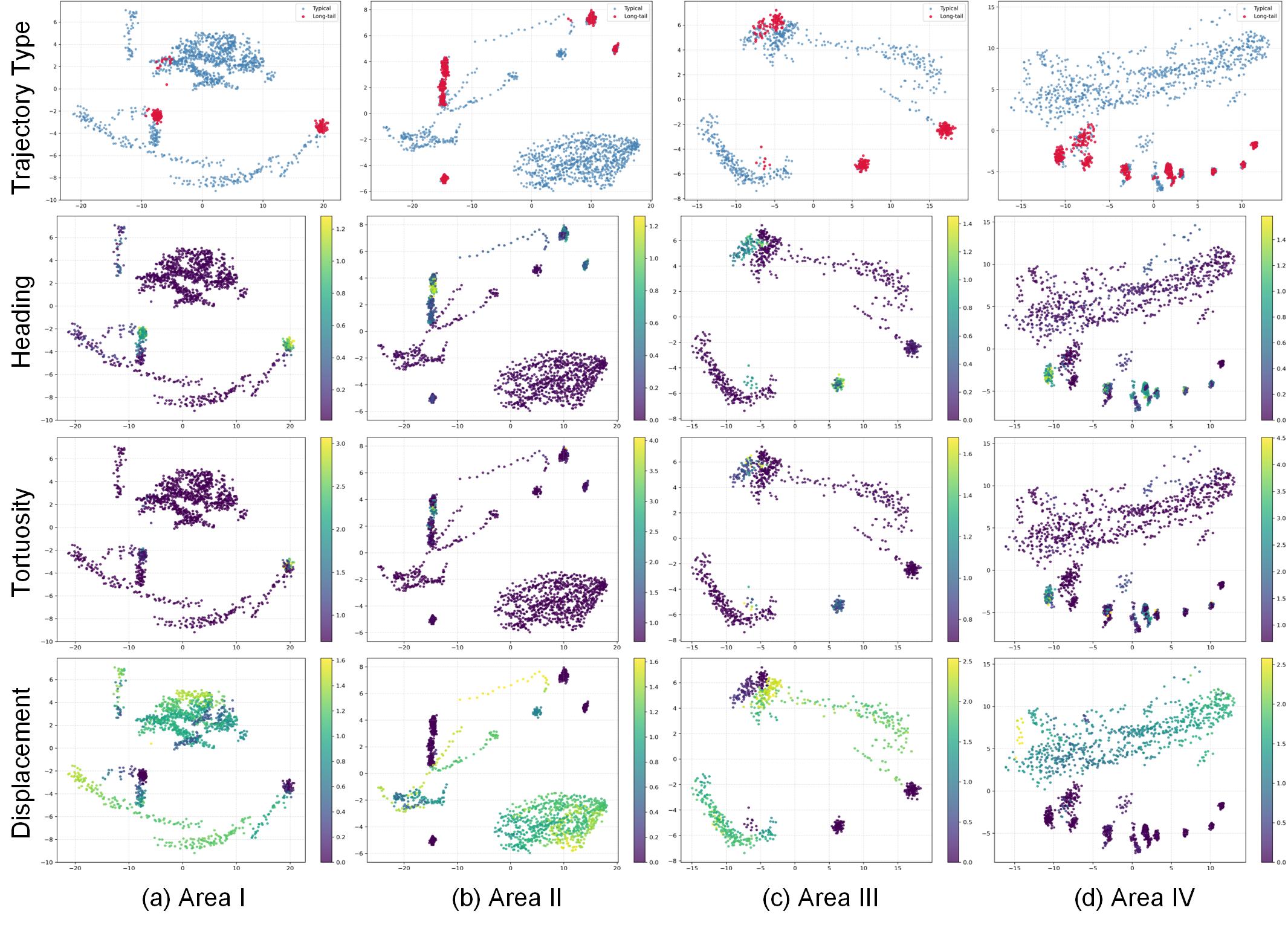}
        \caption{PCA visualization of the learned 2D latent space across four study areas, colored by trajectory type, observed heading change, observed tortuosity, and observed displacement. Long-tail trajectories occupy localized regions, while motion descriptors show structured patterns in the latent space.}
        \label{figure:FU}
    \end{figure}
    \setlength{\tabcolsep}{3.50pt}
    \begin{table}[t]
        \centering
        \caption{Quantitative analysis of multimodal prediction performance on overall, typical, and long-tail trajectories.}
        \begin{tabular}{lcccc}\hline
        Subset & Ratio (\%) & minADE@5 & minFDE@5 & Diversity \\\hline\hline
        Overall               & 100.0 & 0.0823 & 0.1124 & 0.9057 \\
        Typical trajectories  & 84.1  & 0.0795 & 0.1102 & 0.9424 \\
        Long-tail trajectories& 15.9  & 0.0967 & 0.1239 & 0.7118 \\\hline
        \end{tabular}\label{tab:longtail_multimodal}
    \end{table}
    \begin{table}[t]
        \centering
        \caption{Similarity-based analysis of the generalization capability and prototype attraction effect of VGTB. The test set is divided into high- and low-similarity groups according to the similarity between the query trajectory and the retrieved prototype. “B” and “V” denote the base and VGTB-refined predictions, respectively. “Toward GT” indicates the proportion of cases where refinement moves the prediction closer to the ground truth, while “Helpful” and “Harmful” denote beneficial and adverse prototype effects.}
        \setlength{\tabcolsep}{3.5pt}
        \renewcommand{\arraystretch}{1.05}
        \resizebox{\columnwidth}{!}{%
        \begin{tabular}{l|ccccccc}\hline
        Group & B-ADE & V-ADE & B-FDE & V-FDE & Toward GT & Helpful & Harmful \\\hline\hline
        Overall   & 1.016 & 0.148 & 1.072 & 0.250 & 1.000 & 0.864 & 0.000 \\
        High-sim. & 0.854 & 0.084 & 0.986 & 0.155 & 1.000 & 1.000 & 0.000 \\
        Low-sim.  & 1.386 & 0.261 & 1.399 & 0.410 & 1.000 & 0.753 & 0.000 \\\hline
        \end{tabular}\label{tab:vgtb_generalization}
        }
    \end{table}
    \begin{figure}[t]
        \centering
        \setlength{\abovecaptionskip}{-0.25cm}
        \includegraphics[width=1.00\linewidth]{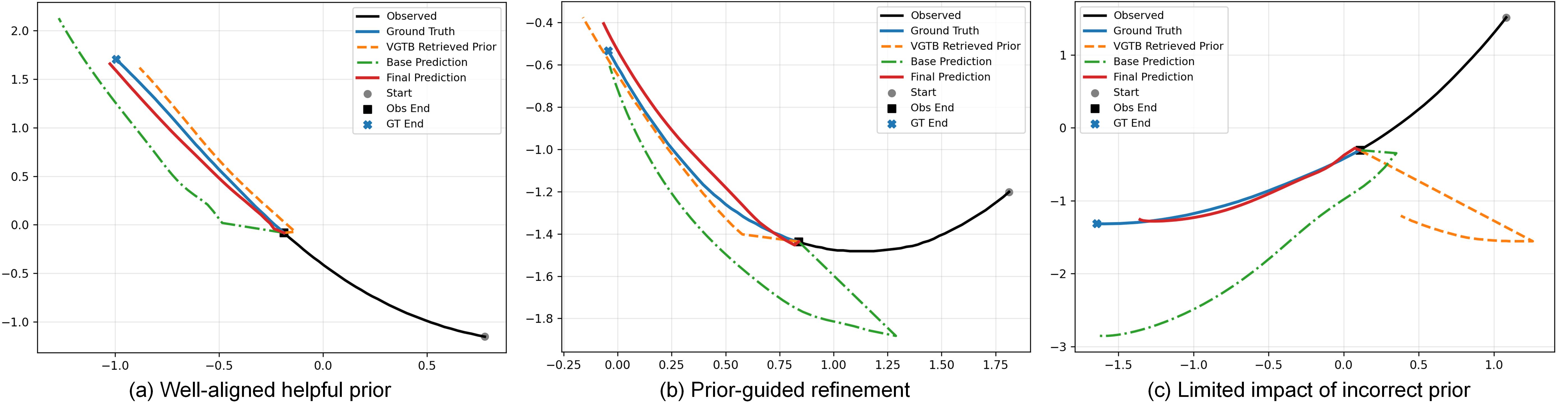}
        \caption{Qualitative visualization of three representative prior-effect cases: (a) well-aligned helpful prior, (b) prior-guided refinement, and (c) limited impact of an incorrect prior on the final prediction.}
        \label{figure:F}
    \end{figure}
    \begin{figure}[t]
        \centering
        \setlength{\abovecaptionskip}{-0.25cm}
        \includegraphics[width=1.00\linewidth]{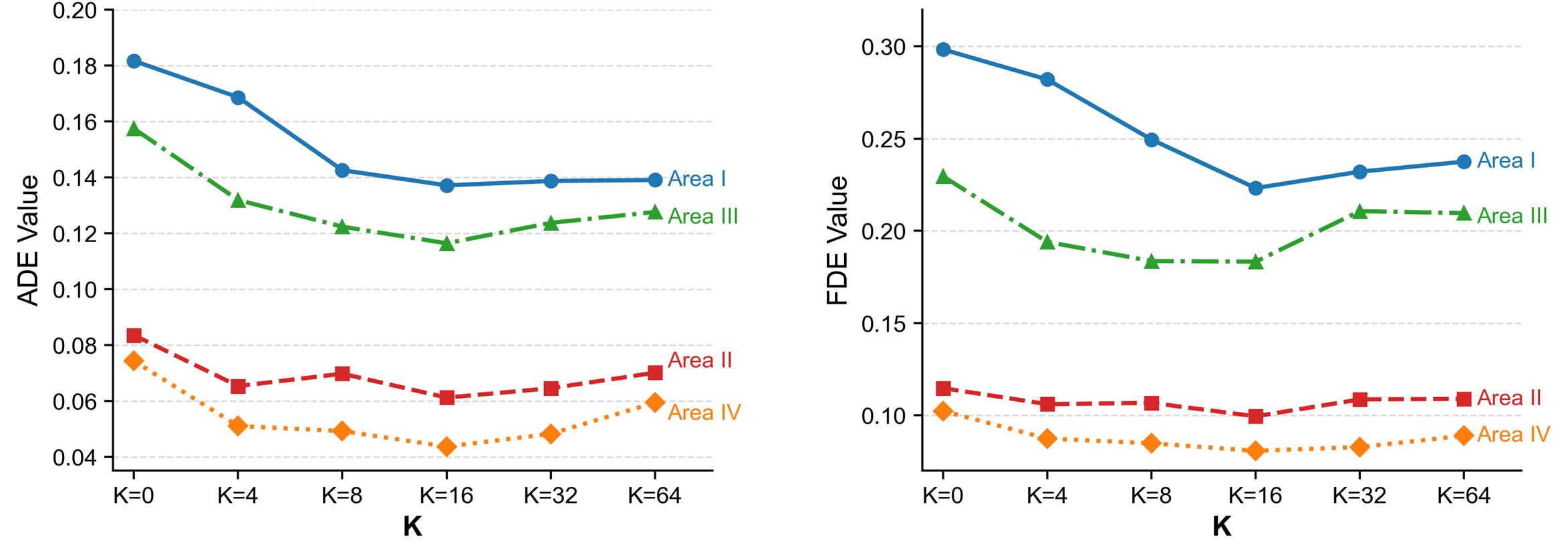}
        \caption{Ablation study on the number of clusters ($K$) in the VGTB module.}
        \label{figure:cluster_ablation}
    \end{figure}
    
    To further investigate the function of the CVAE latent variable, we structurally analyze its learned representations. Although the model operates within a 16-dimensional latent space, we utilize principal component analysis (PCA) to generate 2D projections for visual interpretability. Thus, these visualizations represent low-dimensional approximations of the high-dimensional embeddings rather than the exact latent space. Fig. \ref{figure:FU} shows that the latent codes do not form an unstructured cloud but instead exhibit clear local organization across all four study areas. When colored by trajectory type, long-tail trajectories concentrate in localized regions rather than mixing uniformly with typical ones. Moreover, when the same embeddings are colored by motion descriptors such as observed displacement, heading change, and tortuosity, structured gradients and localized high-value regions emerge. These observations suggest the latent representation correlates with meaningful motion attributes and preserves structured variation associated with both common traffic patterns and rarer maneuvers.
    
    To further verify that the latent space does not merely emphasize average behaviors, we provide a quantitative comparison between typical and long-tail trajectories in Table \ref{tab:longtail_multimodal}. As expected, long-tail cases are more challenging and lead to higher prediction errors than typical ones. Nevertheless, the model still maintains clear multimodal diversity on these rare samples, indicating that the learned latent representation does not collapse to a single averaged prediction. Instead, it remains expressive enough to support multiple plausible futures, including uncommon yet realistic trajectory patterns.
\subsubsection{Analysis of the VGTB}
    To further evaluate the generalization capability of VGTB, we divide the test cases into high-similarity and low-similarity groups based on the similarity between the query trajectory and the retrieved prototype. As shown in Table \ref{tab:vgtb_generalization}, VGTB consistently improves upon the base predictor across all groups, including the low-similarity group, which better reflects challenging cases with rare or less well-represented motion patterns. Moreover, the refined predictions always move toward the ground truth, and no harmful prototype attraction is observed. This indicates that VGTB provides a robust coarse motion prior rather than simply memorizing near-duplicate trajectories. Fig. \ref{figure:F} shows three representative qualitative cases. In Fig. \ref{figure:F}(a), a well-aligned prior directly benefits prediction. In Fig. \ref{figure:F}(b), the prior provides a coarse trend that is further refined by the model. In Fig. \ref{figure:F}(c), even with an inaccurate prior, the final prediction remains largely unaffected and still follows the main ground-truth trend.

    We evaluate $K \in \{0, 4, 8, 16, 32, 64\}$ for VGTB, where $K=0$ removes the module. As shown in Fig. \ref{figure:cluster_ablation}, introducing VGTB significantly reduces prediction errors, and performance improves as $K$ increases to 16 due to more informative trajectory prototypes. Larger values yield only marginal gains with additional computational cost. Therefore, $K=16$ is adopted as a balance between representation capacity and efficiency for our dataset, which involves diverse maneuver patterns in complex waterways; in simpler traffic environments, a smaller $K$ may be sufficient.
\subsection{Discussion and Future Work}
    The proposed CmIVTP effectively addresses the limitations of single-source data in vessel trajectory prediction by integrating multimodal data from AIS and CCTV. The framework demonstrates superior robustness and accuracy in challenging scenarios, including AIS data loss, high vessel density, and complex maritime environments. The innovative components, such as the visual scene target-aware encoder and the cross-modal interaction Transformer, allow the model to capture intricate vessel-environment dynamics, while the uncertainty-aware variational decoder further enhances prediction reliability. These advancements make CmIVTP a promising solution for maritime safety and intelligent decision-making.

    However, some challenges remain. Despite prior efforts to mitigate visual degradation \cite{liu2024real}, extreme environments (e.g., heavy fog, low light) that impair CCTV, or complete AIS loss, still compromise performance. Future research will prioritize enhancing robustness under these conditions and expanding our dataset for real-world deployment. Furthermore, generalizing to new waterways is challenging due to scarce synchronized video data. To address this, we will explore transfer learning by pre-training on abundant AIS data to capture universal motion priors, followed by fine-tuning with limited multimodal data. We also aim to improve the framework's computational efficiency, real-time scalability, and adaptability to diverse sensors. Finally, extending this multimodal fusion to support dynamic traffic scheduling, anomaly detection, and emergency response presents a promising direction for intelligent maritime systems.
\section{Conclusion}
\label{conc}
    This work proposes a novel cross-modal interaction-based vessel trajectory prediction (named CmIVTP) framework that effectively addresses the limitations of single-source data in maritime surveillance. By fusing dynamic AIS data with visual environmental context from CCTV, our method successfully models the intricate interplay between vessel dynamics and environmental constraints. Integrating a visual scene target-aware encoder, a cross-modal interaction Transformer, and an uncertainty-aware variational decoder enables robust trajectory generation that is dynamically consistent and environmentally feasible. Furthermore, a vessel group trajectory bank refines these predictions using historical motion priors, effectively handling complex or rare navigational behaviors. Crucially, the inherent redundancy provided by the multimodal architecture empowers the model to effectively handle AIS data missingness, ensuring continuous and reliable trajectory forecasting in signal-denied environments. Extensive experiments on the newly developed Maritime-MmD$^+$ dataset demonstrate that CmIVTP maintains exceptional stability across varying prediction horizons, high-density traffic, and signal-denied environments, achieving superior accuracy and adaptability compared to better baselines.
\ifCLASSOPTIONcaptionsoff
\newpage
\fi
\bibliographystyle{IEEEtran}
\bibliography{ref.bib}
\tiny

\end{document}